\documentclass[letterpaper]{article} 
\usepackage{aaai23}  
\usepackage{times}  
\usepackage{helvet}  
\usepackage{courier}  
\usepackage[hyphens]{url}  
\usepackage{graphicx} 
\usepackage{natbib}  
\usepackage{caption} 
\usepackage{algorithm}
\usepackage{algorithmic}

\usepackage{amsmath,amsfonts,amsthm,bm}
\usepackage{booktabs}
\usepackage{subfigure}
\usepackage{enumitem}
\usepackage{multirow,color}

\newtheorem{theorem}{Theorem}
\newtheorem{lemma}{Lemma}
\newtheorem{corollary}{Corollary}
\newtheorem{definition}{Definition}
\newcommand{\model}{{{\bf GRADE}}}
\allowdisplaybreaks

\usepackage{newfloat}
\usepackage{listings}
\DeclareCaptionStyle{ruled}{labelfont=normalfont,labelsep=colon,strut=off} 
\floatstyle{ruled}
\newfloat{listing}{tb}{lst}{}
\floatname{listing}{Listing}
\title{Non-IID Transfer Learning on Graphs}
\author{
    Jun Wu\textsuperscript{\rm 1}, ~~Jingrui He\textsuperscript{\rm 1}, ~~Elizabeth Ainsworth\textsuperscript{\rm 1,2}
}
\affiliations{
    \textsuperscript{\rm 1}University of Illinois Urbana-Champaign\\
    \textsuperscript{\rm 2}USDA ARS Global Change and Photosynthesis Research Unit\\

    \{junwu3, jingrui, ainswort\}@illinois.edu
}

\usepackage{bibentry}

\begin{document}

\maketitle

\begin{abstract}
Transfer learning refers to the transfer of knowledge or information from a relevant source domain to a target domain. However, most existing transfer learning theories and algorithms focus on IID tasks, where the source/target samples are assumed to be independent and identically distributed. Very little effort is devoted to theoretically studying the knowledge transferability on non-IID tasks, e.g., cross-network mining.
To bridge the gap, in this paper, we propose rigorous generalization bounds and algorithms for cross-network transfer learning from a source graph to a target graph. The crucial idea is to characterize the cross-network knowledge transferability from the perspective of the Weisfeiler-Lehman graph isomorphism test. To this end, we propose a novel Graph Subtree Discrepancy to measure the graph distribution shift between source and target graphs. Then the generalization error bounds on cross-network transfer learning, including both cross-network node classification and link prediction tasks, can be derived in terms of the source knowledge and the Graph Subtree Discrepancy across domains. This thereby motivates us to propose a generic graph adaptive network (\model) to minimize the distribution shift between source and target graphs for cross-network transfer learning. Experimental results verify the effectiveness and efficiency of our \model\ framework on both cross-network node classification and cross-domain recommendation tasks.
\end{abstract}

\section{Introduction}
Transfer learning~\cite{pan2009survey} tackles the knowledge transferability from a source domain to a relevant target domain under a distribution shift. It has been theoretically shown~\cite{ben2010theory,zhang2019bridging,pmlr-v139-acuna21a} that the generalization performance of a learning algorithm can be improved by leveraging the knowledge from the source domain, when the source and target domains have the same labeling space (also known as domain adaption~\cite{pan2009survey,zhao2019learning}). To be more specific, the expected target error could be bounded in terms of the prediction error of the source domain and the distribution discrepancy across domains. It thus motivates a line of practical approaches with domain discrepancy minimization in the latent feature space~\cite{ganin2016domain,pmlr-v139-acuna21a}. However, it is noteworthy that most of the existing theoretical guarantees and empirical algorithms hold the IID assumption that all the source/target samples are drawn independently from an identical source/target distribution. This hinders their applications on other tasks with non-IID data, e.g., node classification~\cite{kipf2017semi,wu2019scalable,tang2020investigating} and recommendation~\cite{zhao2019cross,zhou2021intrinsic,zhou2021pure} across domains. 

\begin{figure}[!t]
    \centering
    \includegraphics[width=0.47\textwidth]{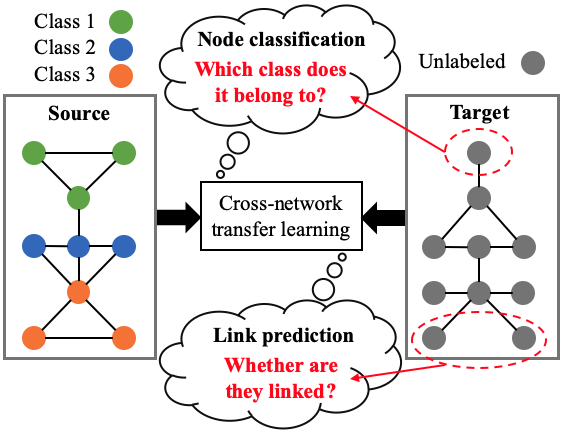}
    \caption{Illustration of the cross-network transfer learning (best viewed in color). Given a labeled source graph (color indicates node label) and an unlabeled target graph, cross-network transfer learning tackles the classification and link prediction tasks in the target graph, by leveraging the auxiliary information from the source graph.}
    \label{fig:CNTL}
\end{figure}

\begin{figure}[!t]
    \centering
    \includegraphics[width=0.47\textwidth]{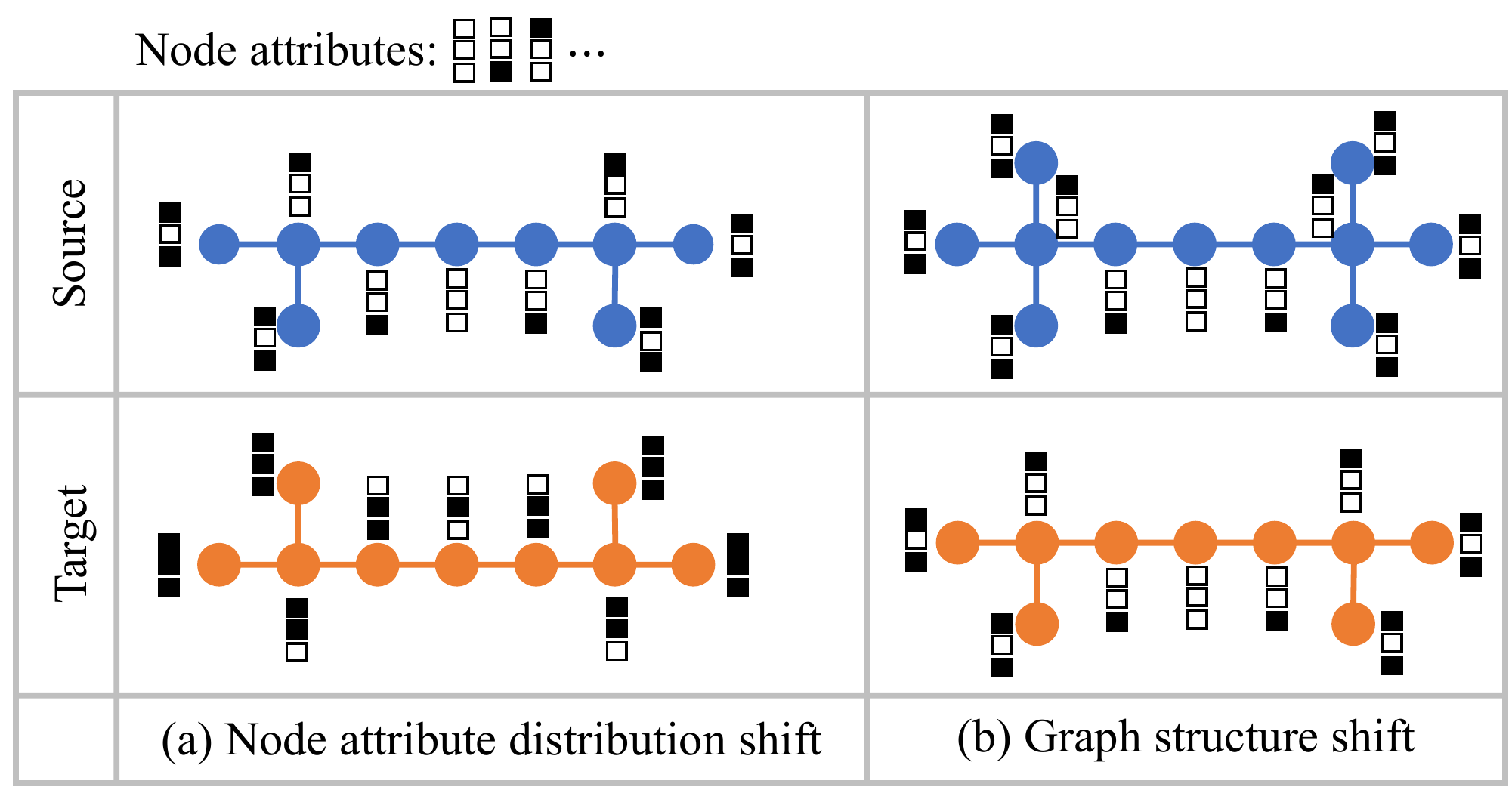}
    \caption{Illustration of distribution shift on graphs. (a) Source and target graphs share the same graph structure, but they have different node attribute distributions (node attribute is assumed to be a 3-dimensional feature vector, where the ``black" box denotes 1 and the ``white" box denotes 0). (b) Source and target graphs share a similar node attribute distribution but different graph structures.}
    \label{fig:distribution_shift}
\end{figure}

In this paper, we focus on studying the problem of cross-network transfer learning, where the knowledge can be transferred from a source graph to a target graph\footnote{In this paper, we use ``graph" and ``network" interchangeably to denote the graph-structured data in every domain.}. To be specific, we consider the following cross-network mining tasks (see Figure~\ref{fig:CNTL}). (1) Node classification (e.g., cross-network role identification~\cite{zhu2021transfer}): It aims to predict the class labels of nodes, by leveraging the knowledge from a source graph with fully labeled nodes. Following~\cite{wu2020unsupervised}, we consider the unsupervised learning scenarios where the target domain has no label information.  (2) Link prediction (e.g., cross-domain recommendation~\cite{li2020ddtcdr,zhao2019cross}): It predicts the missing links in the incomplete target graph, by leveraging knowledge from a complete source graph. The unique challenge of cross-network transfer learning lies in the interdependence nature of nodes within the graph. As shown in Figure~\ref{fig:distribution_shift}, the distribution shift between source and target graphs can be induced by node attribute and graph structure.

We start by developing a novel distribution discrepancy measure named Graph Subtree Discrepancy between source and target graphs. This is motivated by the connection of existing message-passing graph neural networks~\cite{hamilton2017inductive,xu2018representation,xu2018how} and Weisfeiler-Lehman graph kernel~\cite{weisfeiler1968reduction,shervashidze2011weisfeiler}. On one hand, the Weisfeiler-Lehman graph kernel holds that the non-parametric graph similarity can be decomposed into the similarity of a sequence of subtrees rooted at every node. On the other hand, message-passing graph neural networks tend to iteratively aggregate the messages from nodes' local neighborhoods in a parametric way. Then, Graph Subtree Discrepancy is designed to measure the distribution shift of graphs by estimating the similarity/difference of subtree representations learned from a message-passing graph neural network. As a result, it can inherit the benefits of high expressiveness from message-passing graph neural networks and feasible explanations from the Weisfeiler-Lehman graph kernel. Based on Graph Subtree Discrepancy, the generalization error bounds of cross-network transfer learning can be derived for cross-network mining tasks. By empirically minimizing the error upper bounds, we propose a generic graph adaptive network (\model) for cross-network transfer learning. The efficacy of the proposed \model\ framework is confirmed on various cross-network mining data sets. The major contributions of this paper are summarized as follows.
\begin{itemize}[leftmargin=*,noitemsep,nolistsep]
    \item We propose a novel Graph Subtree Discrepancy to measure the distribution shift of nodes' data distribution between source and target graphs. The generalization error bounds of cross-network transfer learning can then be derived based on Graph Subtree Discrepancy.
    \item We propose a generic Graph Adaptive Network (\model) framework for cross-network transfer learning, followed by the instantiations on cross-network node classification and cross-domain recommendation tasks.
    \item Extensive experiments demonstrate the effectiveness of our proposed \model\ framework on cross-network node classification and cross-domain recommendation tasks.
\end{itemize}


\section{Related Work}\label{sec:related_work}

\subsection{Transfer Learning}
Transfer learning~\cite{pan2009survey} refers to the knowledge transferability from a source domain to a relevant target domain. It is theoretically guaranteed~\cite{mansour2009domain,ben2010theory,pmlr-v139-acuna21a,wu2021indirect,wu2022domain,wu2022unified,wu2022distributioninformed,wu2020continuous} that under mild conditions, the generalization performance of a learning algorithm on the target domain can be improved by leveraging the knowledge from the source domain. One common assumption behind those theoretical guarantees is that all the source/target samples are drawn independently from an independent and identical source/target probability distribution. More recently, \cite{levie2019transferability,ruiz2020graphon,zhu2021transfer} proposed to analyze the transferability of graph neural networks using graphons or ego-graphs. Nevertheless, those works explore whether graph neural networks are transferable given two graphs. In contrast, in this paper, by using a hypothesis-dependent Graph Subtree Discrepancy, we show how knowledge can be transferred across graphs. The resulting theoretical analysis provides insight into designing practical cross-network transfer learning algorithms.

\subsection{Cross-Network Mining}
Cross-network mining aims to exhibit the informative patterns from multiple relevant networks/graphs for a variety of mining tasks, e.g. cross-network node classification~\cite{wu2020unsupervised,zhang2019dane,zhu2021transfer}, multi-domain graph clustering~\cite{ni2015flexible}, cross-domain recommendation~\cite{zhao2019cross,li2020ddtcdr}, etc. There are two lines of solutions in exploring the knowledge transferability among different graphs. One is~\cite{fang2015trgraph,wu2020unsupervised,zhang2019dane,dai2022graph} that it extracts the signature subgraphs or consistent aggregation patterns from source and target graphs without a theoretical explanation. The other one is~\cite{hu2019strategies,hu2020gpt,qiu2020gcc,han2021adaptive} that it first pre-trains the graph neural networks on a large source graph for encoding the general graph structures, and then fine-tunes on the target graph for extracting the task-specific information. This might lead to a sub-optimal solution for unsupervised cross-network node classification tasks where no labeled target nodes are available for fine-tuning. 

\section{Preliminaries}\label{sec:Preliminaries}

\subsection{Notation}
Suppose that a graph is represented as $G=(V, E)$, where $V=\{v_1,\cdots,v_{n}\}$ is the set of $n$ nodes and $E \subseteq V \times V$ is the edge set in the graph. In this paper, we consider the attributed graph. That is, each node is associated with a $D$-dimensional feature vector $x_v\in \mathbb{R}^D$. In the node classification task, each node is associated with a class label $y_v \in \{1,\cdots, C\}$, where $C$ is the total number of classes. The graph $G$ can also be represented by an adjacency matrix $A\in \mathbb{R}^{n\times n}$, where $A_{ij}$ is the similarity between $v_i$ and $v_j$ on the graph. In the context of cross-network network mining, we denote $G^s=(V^s,E^s,X^s)$ and $G^t=(V^t,E^t,X^t)$ to be the source and target graphs, respectively. The associated adjacent matrices of source and target graphs are represented as $A^s$ and $A^t$, respectively.

\subsection{Problem Setting}
Following~\cite{wu2020unsupervised,zhu2021transfer}, we formally define the cross-network transfer learning problem as follows.

\begin{definition}({\bf Cross-Network Transfer Learning})
Given a source graph $G^s$ and a target graph $G^t$, cross-network transfer learning aims to improve the prediction performance of node classification or link prediction in the target graph by using knowledge from the source graph, with the assumption that source and target graphs are related.
\end{definition}

As illustrated in Figure~\ref{fig:distribution_shift}, the distribution shift across graphs can be induced by both node attribute\footnote{In this paper, we consider that source and target graphs share the same node attribute space, but they might have different node attribute distributions.} and graph structure. Compared to standard transfer learning~\cite{ben2010theory}, the additional distribution shift over complex graph structure leads to a much more challenging cross-network transfer learning problem setting.

\section{Theoretical Results}\label{sec:theory}
In this section, we propose a novel Graph Subtree Discrepancy (GSD) to measure the distribution shift across graphs. 

\subsection{Connection of WL Kernels and GNNs}
Weisfeiler-Lehman graph subtree kernel~\cite{shervashidze2011weisfeiler} aims to measure the semantic similarity of a pair of input graphs.
It learns a sequence of Weisfeiler-Lehman subgraphs for an input graph $G$: $\{G_0, G_1, \cdots, G_m, \cdots\} = \{(V,E,f_0),(V,E,f_1),$ $\cdots,(V,E,f_m),\cdots\}$,
where $G_0=G$ and $f_0(v)$ denotes the raw node attributes of $v$ for any $v\in G$. The ``relabeling'' function $f_j$ ($j=1,\cdots,m,\cdots$) aims to represent the subtree structure rooted at $v\in G$ into a novel representation at every iteration (see Definition~\ref{def:subtree}). Then, the structural information around node $v$ can be represented as a sequence of Weisfeiler-Lehman subtrees $\{f_0(v), f_1(v), \cdots, f_m(v), \cdots\}$ with different depths $m$.

\begin{definition}\label{def:subtree} (Weisfeiler-Lehman Subtree~\cite{shervashidze2011weisfeiler}) Given a graph $G=(V,E)$ associated with initial node attributes $f_0(v)$ for $v\in V$, the Weisfeiler-Lehman subtree of depth $m$ rooted at $v\in V$  can be represented as $f_m(v):=f_m\left(f_{m-1}(v); \cup_{u\in N(v)} f_{m-1}(u) \right)$ where $N(v)$ denotes the nearest neighbors of root node $v$.
\end{definition}
Note that in the original work~\cite{shervashidze2011weisfeiler}, the ``relabeling'' function of the WL subtree is simply given by the hashing table due to the discrete node attributes in the graph. Later, it is revealed~\cite{hamilton2017inductive,demonet_kdd19,geerts2021let} that this WL subtree can actually recover the crucial message-passing modular in many popular message-passing GNNs, where the ``relabeling'' function $f_m(\cdot)$ is instantiated by deep neural networks for learning continuous node representation. We observe that for single graph mining task, e.g., node classification and link prediction, only the message-passing philosophy of the Weisfeiler-Lehman subtree is studied to design the graph neural networks~\cite{hamilton2017inductive,kipf2017semi}. It maps the structurally equivalent nodes within one graph into the same low-dimensional representation in a latent feature space. However, in the context of cross-network transfer learning, we highlight that the following WL subtree kernel sheds light on measuring the distribution shift of source and target graphs in the feature space learned by GNNs.

\begin{definition}\label{def:WL_kernel} (Weisfeiler-Lehman Subtree Kernel~\cite{shervashidze2011weisfeiler}) Given any two graphs $G=(V,E)$ with $n$ nodes and $G'=(V',E')$ with $n'$ nodes, the Weisfeiler-Lehman subtree kernel on two graphs $G$ and $G'$ with $M$ iterations is defined as:
\begin{align*}
    k\left(G, G'\right) = \frac{1}{nn'} \sum_{m=0}^M \sum_{v\in G}\sum_{v' \in G'} \delta\left( f_m(v), f_m(v') \right)
\end{align*}
where $\delta(\cdot, \cdot)$ is the Dirac kernel, that is, it is 1 when its arguments are equal and 0 otherwise, and $f_m(v)$ represents the subtree pattern of depth $m$ rooted at $v$.
\end{definition}
The WL subtree kernel on graphs $G$ and $G'$ is rewritten as
\begin{align*}
    k\left(G, G'\right) = \sum_{m=0}^M s\left( \hat{\mathbb{P}}\left(G_m\right), \hat{\mathbb{Q}}\left(G'_m\right) \right)
\end{align*}
where $\hat{\mathbb{P}}$ (or $\hat{\mathbb{Q}}$) is the empirical node distribution of graph $G$ (or $G'$), i.e., $\hat{\mathbb{P}}(\tau|G_m) = \frac{1}{n}\sum_{i=1}^{n}\delta(f_m(v_i), \tau)$ for any subtree pattern $\tau$. Here $s(\cdot, \cdot)$ is an inner product metric to measure the distribution similarity of $\hat{\mathbb{P}}(G_m)$ and $\hat{\mathbb{Q}}(G'_m)$, i.e., $s(\hat{\mathbb{P}}(G_m), \hat{\mathbb{Q}}(G'_m))= \langle (\hat{\mathbb{P}}(\tau_1|G_m), \cdots,$ $ \hat{\mathbb{P}}(\tau_k|G_m), \cdots), (\hat{\mathbb{P}}(\tau_1|G'_m), \cdots, \hat{\mathbb{P}}(\tau_k|G'_m), \cdots) \rangle$. Furthermore, we have the following observations. (1) This nonparametric graph kernel~\cite{shervashidze2011weisfeiler} can be exploited to measure the distribution shift between source and target graphs by counting the occurrence of subtrees when the node attributes are discrete. However, it might suffer when using continuous node attribute to estimate the graph similarity (or distribution discrepancy in the context of cross-network transfer learning). (2) Previous works~\cite{wu2020unsupervised,zhu2021transfer,dai2022graph} focus on characterizing the distribution discrepancy over only the $m^{\text{th}}$ subtree representation. They can thereby partially reveal the distribution discrepancy between source and target graphs in practice.

\subsection{Graph Subtree Discrepancy}
Inspired by the connection of the WL subtree kernel and message-passing GNNs, we propose a parametric Graph Subtree Discrepancy (GSD). GSD measures the distribution discrepancy of graphs in the latent feature space induced by the message-passing GNNs.

Following WL subtree kernel~\cite{shervashidze2011weisfeiler}, we assume that given a graph $G=(V, E)$, the WL subtrees (i.e., $\{f_m(v) | v\in V\}$) with a fixed depth $m$ are conditionally independent with respect to WL subgraph $G_{m-1}$ at depth $m-1$, i.e., $f_m (u) \perp f_m(v) | G_{m-1}$. In this case, given the WL subgraph $G_{m-1}$, the subtree representations $\{f_m(v) | v\in V\}$ can thus be considered as IID samples. This tells us that the subtree (at depth $m$) distribution shift of source and target graphs can be measured by any existing distribution discrepancy measures, e.g, $JS$-divergence~\cite{ben2010theory,ganin2016domain}, discrepancy distance~\cite{mansour2009domain}, Maximum Mean Discrepancy~\cite{gretton2012kernel} and $f$-divergence~\cite{pmlr-v139-acuna21a}. Therefore, our Graph Subtree Discrepancy can be formally defined as follows.
\begin{definition} (Graph Subtree Discrepancy)
Given two graphs $G^s=(V^s,E^s)$ and $G^t=(V^t,E^t)$, the graph subtree discrepancy between them can be defined as:
\begin{equation}
    d_{GSD}\left(G^s, G^t\right) = \lim_{M\to \infty} \frac{1}{M+1} \sum_{m=0}^M d_b\left(G_m^s, G_m^t\right)
\end{equation}
where $d_b(\cdot,\cdot)$ is the base domain discrepancy.
\end{definition}
For example, we can use the discrepancy distance~\cite{mansour2009domain} to instantiate the base domain discrepancy $d_b(\cdot,\cdot)$, which is defined as
\begin{align}\label{eq:discrepancy_distance}
    d_b(G^s_m, G^t_m) = \sup_{h, h'\in \mathcal{H}} \big| \mathbb{E}_{v\in V^s}\left[ L\left(h\left(f_m(v)), h'(f_m(v)\right)\right) \right] \nonumber \\ 
    - \mathbb{E}_{v\in V^t}\left[ L\left(h\left(f_m(v)), h'(f_m(v)\right)\right) \right] \big|
\end{align}

We see that GSD recursively estimates the subtrees' distribution discrepancy between source and target graphs at different depths. Here the subtree representation can be learned by existing passage-passing GNNs~\cite{hamilton2017inductive,velivckovic2018graph,xu2018how}.

\subsection{Error Bounds}
Next, we derive the error bounds for cross-network transfer learning based on GSD. Let $\mathcal{H}$ be the hypothesis space. For any hypothesis $h\in \mathcal{H}$, the node classification error on graph $G$ of a message-passing GNN (with $L$ convolutional layers) can be defined as $\epsilon(h\circ f) = \mathbb{E}_{v\in G}[\mathcal{L}(h(f(v)), y)]$, where $f(\cdot)$ extracts the node representation and $\mathcal{L}(\cdot, \cdot)$ is the loss function. Without loss of generality, we focus on the commonly used GNNs with the feature extraction $f(v) = f_L(v)$ (using only the output of the final graph convolutional layer~\cite{kipf2017semi,hamilton2017inductive,velivckovic2018graph}). The following theorem shows that in the cross-network node classification, the expected error in the target graph can be bounded in terms of the classification error in the source graph and the distribution discrepancy across graphs.

\begin{theorem}\label{thm:TL_node_classification} (Cross-Network Node Classification)
Assume that there are a source graph $G^s$ and a target graph $G^t$ and the base domain discrepancy $d_b(\cdot,\cdot)$ of GSD is instantiated by the discrepancy distance (see Eq. (\ref{eq:discrepancy_distance})). Given a message-passing GNN with the feature extractor $f$ and the hypothesis $h\in \mathcal{H}$, the node classification error in the target graph can be bounded as follows.
\begin{equation*}
    \epsilon_t(h\circ f) \leq \epsilon_s(h\circ f) + d_{GSD}\left(G^s, G^t\right) + \lambda^* + R^*
\end{equation*}
where $\lambda^* = \mathbb{E}_{v\in V^t}[\mathcal{L}(h^s_*(f(v)), h^t_*(f(v)))]$ measures the prediction difference of optimal source and target hypotheses on the target nodes, and $R^*=\mathbb{E}_{v\in V^s}[\mathcal{L}(y, h^s_*(f(v)))] + \mathbb{E}_{v\in V^t}[\mathcal{L}(h^t_*(f(v)), y)]$ is the Bayes error on the source and target graphs. $y$ is the class label of $v$. In this case, $h^s_* \in \arg\min_{h\in \mathcal{H}}$ $\mathbb{E}_{v\in V^s}[\mathcal{L}(h(f(v)), y)]$ and $h^t_* \in \arg\min_{h\in \mathcal{H}} \mathbb{E}_{v\in V^t}[\mathcal{L}(h(f(v)), y)]$ are the optimal source and target hypotheses, respectively.
\end{theorem}

Compared to previous work~\cite{zhu2021transfer}, our error bound of Theorem~\ref{thm:TL_node_classification} is hypothesis-dependent. That is, the knowledge transferability can be enhanced, if the message-passing GNN learns a latent feature space such that the subtree distribution shift of source and target graphs is minimized. This is in sharp contrast to previous work which focuses on evaluating the transferability of a trained GNN model. Therefore, Theorem~\ref{thm:TL_node_classification} provides insights on designing practical cross-network transfer learning algorithms by minimizing the error upper bound.

We have similar results for cross-network link prediction. That is, the label space of link prediction is $\mathcal{Y}=\{0,1\}$, where $y=1$ if the link of a pair of nodes exists, $y=0$ otherwise. In this case, the loss of the intra-graph link prediction is defined as $\epsilon(h\circ f) = \mathbb{E}_{u,v\in V\times V}[\mathcal{L}(h([f_L(u) || f_L(v)]), y)]$, where $[\cdot || \cdot]$ denote the vector concatenation.
\begin{theorem}\label{thm:TL_link_prediction} (Cross-Network Link Prediction)
With assumptions in Theorem~\ref{thm:TL_node_classification}, and let $\mathcal{L}(y,\Tilde{y}) = |y-\Tilde{y}|$ and the hypothesis class $\mathcal{H}$ is given by the multi-layer perceptrons, if the loss of the link prediction is defined as $\epsilon^{link}(h\circ f) = \mathbb{E}_{u,v\in V\times V}[\mathcal{L}(h([f_L(u) || f_L(v)]), y)]$, then the link prediction error in the target graph can be bounded as follows.
\begin{align*}
    \epsilon_t^{link}(h) \leq \epsilon_t^{link}(h) + d_{GSD}\left(G^s, G^t\right) + \lambda^*_{link} + R^*_{link}
\end{align*}
where $\lambda^*_{link} = \mathbb{E}_{(u,v)\in V^t\times V^t}[\mathcal{L}(h^s_*([f(u) || f(v)]),$ $ h^t_*([f(u) || f(v)]))]$ measures the difference of optimal source and target hypotheses on the target graph, and $R^*_{link}=\mathbb{E}_{(u,v)\in V^s\times V^s}[\mathcal{L}(y, h^s_*([f(u) || f(v)]))] + \mathbb{E}_{(u,v)\in V^t\times V^t}[\mathcal{L}(h^t_*([f(u) || f(v)]), y)]$ is the Bayes error. In this case, $h^s_* \in \arg\min_{h\in \mathcal{H}}$ $\mathbb{E}_{(u,v)\in V^s\times V^s}[\mathcal{L}(h([f(u) || f(v)]), y)]$, and $h^t_* \in \arg\min_{h\in \mathcal{H}} \mathbb{E}_{(u,v)\in V^t\times V^t}[\mathcal{L}(h([f(u) || f(v)]), y)]$ are optimal source and target hypothesises, respectively.
\end{theorem}


\section{Proposed Framework}\label{sec:algorithm}
In this section, we propose a cross-network transfer learning framework named \underline{Gr}aph \underline{Ad}aptive N\underline{e}twork (\model).

\subsection{Objective Function}
The objective function of a generic cross-network transfer learning framework (\model) is summarized as follows.
\begin{align}\label{eq:framework}
    \min_{\theta} C(G^s; \theta) + \lambda \cdot d_{GSD}(G^s, G^t; \theta)
\end{align}
where $\theta$ denotes all the trainable parameters. $C(G^s; \theta)$ is the task-specific loss function on the source graph, and $d_{GSD}(G^s, G^t; \theta)$ is the discrepancy minimization between source and target graphs. $\lambda \geq 0$ is a hyper-parameter to balance these terms. Note that $C(G^s; \theta)$ might also contain the task-specific loss function on the target graph, if label information is partially available in the target domain.

\subsection{Algorithms}
Following the framework of Eq. (\ref{eq:framework}), we present the instantiated algorithms for two cross-network mining tasks, including cross-network node classification ({\bf \model-N}) and cross-domain recommendation ({\bf \model-R}).

\subsubsection{Cross-Network Node Classification}
We focus on the cross-network node classification setting from a source graph with labeled nodes to a target graph with only unlabeled nodes. The goal is to identify the class label of every node in the target domain, by leveraging the relevant knowledge from the source domain. The objective function of {\bf \model-N} can be instantiated as follows.
\begin{equation}\label{eq:grade_nc}
    \begin{aligned}
        \min_{\theta_f, \theta_h} &\quad \mathcal{L}\left( h\left(f(G^s; \theta_f); \theta_h\right), Y^s \right) \\
        &\quad + \lambda \cdot d_{GSD}\left( f(G^s; \theta_f), f(G^t; \theta_f) \right)
    \end{aligned}
\end{equation}
where $f(\cdot)$ is the message-passing graph neural network function parameterized by $\theta_f$, and $h(\cdot)$ is the classifier function (MLP is adopted in the experiments) parameterized by $\theta_h$. $\mathcal{L}(\cdot, \cdot)$ is the cross-entropy loss function for cross-network node classification in the experiments (mean square error loss function can be applied for regression task).

Specifically, we adopt Graph Convolutional Network (GCN)~\cite{kipf2017semi} as the base model to extract the subtree representations of a graph. Then, the subtree pattern of depth $m$ rooted at $v$ can be represented as follows.
\begin{equation}\label{eq:gcn}
f_m(v) = \sigma \Big(\sum\nolimits_{u \in \{v\} \cup N(v)} \widehat{a}_{vu} W^{m} f_{m-1}(u) \Big)
\end{equation}
where $\widehat{A} = (\widehat{a}_{vu}) \in \mathbb{R}^{n\times n}$ ($n$ is the number of nodes) is the re-normalization of the adjacency matrix $A$ with added self-loops, and $W^{m}$ is the trainable matrix at $m^{\textrm{th}}$ layer. $\sigma(\cdot)$ is the non-linear activation function. After $M$ iterations, we obtain the sequence of subtree representations rooted at $v$: $f_0(v), f_1(v),\cdots, f_M(v)$, where $f_0(v)$ is the raw node attributes of $v$. Following~\cite{kipf2017semi}, the final representation $f_M(v)$ can be applied to identify the class label of node $v$.
In addition, we consider finite iterations (e.g., $M$) of the message-passing graph neural network for estimating GSD. That is,
\begin{align}\label{eq:practical_gsd}
    d_{GSD}\left( f(G^s; \theta_f), f(G^t; \theta_f) \right) = \frac{1}{M+1} \sum_{m=0}^M d_b\left(G_m^s, G_m^t\right)
\end{align}

\subsubsection{Cross-Domain Recommendation}
Cross-domain recommendation learns the user preference in the target domain, by leveraging the rich information from a relevant source domain. The objective function of {\bf \model-R} can be instantiated as follows.
\begin{equation}
    \begin{aligned}
    \min_{\theta_f, \theta_{h'}} &\quad \mathcal{L} \left( h'\left(f(G^s; \theta_f); \theta_{h'}\right) \right) + \mathcal{L} \left( h'\left(f(G^t; \theta_f); \theta_{h'}\right) \right) \\
    &\quad + \lambda \cdot d_{GSD}\left( f(G^s; \theta_f), f(G^t; \theta_f) \right)
\end{aligned}
\end{equation}
where $f(\cdot)$ is the message-passing graph neural network function parameterized by $\theta_f$, and $h'(\cdot)$ is the link prediction function parameterized by $\theta_{h'}$.

More specifically, we adopt GCN (see Eq. (\ref{eq:gcn})) as the base model $f(\cdot)$ to extract the subtree representations of a graph. The graph subtree discrepancy $d_{GSD}(\cdot, \cdot)$ can also be given by Eq. (\ref{eq:practical_gsd}) over those subtree representations. In addition, following~\cite{he2017neural,zhao2019cross}, we adopt the multi-layer perceptron as the link prediction function $h'(\cdot)$ to infer whether a link of two nodes exists. That is, 
\begin{align*}
    h' \left( (u, v), y_{uv} ; \theta_{h'} \right) = \text{BCE} \left( \text{MLP}_{\theta_{h'}}\left(f(u) || f(v) \right), y_{uv} \right)
\end{align*}
where $y_{uv}$ is the link label (i.e., $y_{uv}=1$ if $u$ and $v$ are linked, $y_{uv}=0$ otherwise) for any $u,v \in V^s$ or $u,v \in V^t$. Here $\text{BCE}(\cdot)$ denotes the binary cross-entropy, and $\text{MLP}_{\theta_{h'}}(\cdot)$ is a multi-layer perceptron function.

\section{Experiment}\label{sec:experiments}
\begin{table*}[!t]
\centering
\small
\begin{tabular}{|l|cccccc|c|}
\toprule
Methods & USA $\rightarrow$ Brazil & USA $\rightarrow$ Europe & Brazil $\rightarrow$ USA & Brazil $\rightarrow$ Europe & Europe $\rightarrow$ USA & Europe $\rightarrow$ Brazil & Avg. \\\midrule
GCN & 0.366$_{\pm \text{0.011}}$ &	0.371$_{\pm \text{0.004}}$ &	0.491$_{\pm \text{0.011}}$ &	0.452$_{\pm \text{0.012}}$ &	0.439$_{\pm \text{0.001}}$ &	0.298$_{\pm \text{0.022}}$ & 0.403 \\
SGC & 0.527$_{\pm \text{0.022}}$ &	0.430$_{\pm \text{0.009}}$ &	0.432$_{\pm \text{0.005}}$ &	0.479$_{\pm \text{0.000}}$ &	0.447$_{\pm \text{0.002}}$ &	0.481$_{\pm \text{0.011}}$ & 0.466 \\
GCNII & 0.344$_{\pm \text{0.086}}$ &	0.393$_{\pm \text{0.025}}$ &	0.470$_{\pm \text{0.056}}$ &	0.494$_{\pm \text{0.018}}$ &	0.460$_{\pm \text{0.012}}$ &	0.542$_{\pm \text{0.011}}$ & 0.450 \\ \midrule
DAN & 0.504$_{\pm \text{0.020}}$ &	0.393$_{\pm \text{0.000}}$ &	0.436$_{\pm \text{0.006}}$ &	0.393$_{\pm \text{0.010}}$ &	0.436$_{\pm \text{0.003}}$ &	0.542$_{\pm \text{0.000}}$ & 0.451 \\
DANN & 0.500$_{\pm \text{0.005}}$ &	0.386$_{\pm \text{0.011}}$ &	0.402$_{\pm \text{0.048}}$ &	0.350$_{\pm \text{0.062}}$ &	0.436$_{\pm \text{0.000}}$ &	0.538$_{\pm \text{0.005}}$ & 0.435 \\
MDD & 0.500$_{\pm \text{0.005}}$ &	0.378$_{\pm \text{0.000}}$ &	0.402$_{\pm \text{0.048}}$ &	0.350$_{\pm \text{0.062}}$ &	0.402$_{\pm \text{0.048}}$ &	0.477$_{\pm \text{0.081}}$ & 0.418 \\ \midrule
AdaGCN & 0.466$_{\pm \text{0.065}}$ &	0.434$_{\pm \text{0.004}}$ &\bf	0.501$_{\pm \text{0.003}}$ &	0.486$_{\pm \text{0.021}}$ &	0.456$_{\pm \text{0.034}}$ &	0.561$_{\pm \text{0.081}}$ & 0.484 \\
UDA-GCN &\bf 0.607$_{\pm \text{0.059}}$ &	0.388$_{\pm \text{0.007}}$ &	0.497$_{\pm \text{0.005}}$ &\bf	0.510$_{\pm \text{0.019}}$ &	0.434$_{\pm \text{0.042}}$ &	0.477$_{\pm \text{0.024}}$ & 0.486 \\
EGI & 0.523$_{\pm \text{0.013}}$ &	0.451$_{\pm \text{0.011}}$ &	0.417$_{\pm \text{0.021}}$ &	0.454$_{\pm \text{0.046}}$ &	0.452$_{\pm \text{0.029}}$ &\bf	0.588$_{\pm \text{0.011}}$ & 0.481 \\ \midrule
{\bf \model-N} & 0.550$_{\pm \text{0.062}}$ &\bf	0.457$_{\pm \text{0.027}}$ &	0.497$_{\pm \text{0.010}}$ &	0.506$_{\pm \text{0.004}}$ &\bf	0.463$_{\pm \text{0.001}}$ &\bf	0.588$_{\pm \text{0.032}}$ &\bf 0.510 \\
\bottomrule
\end{tabular}
\caption{Cross-network node classification on the Airport network}
\label{tab:airport_classification}
\end{table*}

\begin{table}[!t]
    \centering
    \small
    \setlength\tabcolsep{3.7pt}
    \begin{tabular}{|l|ccccc|}
    \toprule
        \multirow{2}{*}{Methods} & \multicolumn{2}{c}{Citation} & & \multicolumn{2}{c|}{Social} \\ \cline{2-3}\cline{5-6}
         & A $\rightarrow$ D & D $\rightarrow$ A && B1 $\rightarrow$ B2 & B2 $\rightarrow$ B1 \\\midrule
        GCN & 0.435$_{\pm \text{0.013}}$ & 0.567$_{\pm \text{0.046}}$  & & 0.408$_{\pm \text{0.025}}$  & 0.451$_{\pm \text{0.044}}$ \\
        SGC & 0.430$_{\pm \text{0.001}}$ & 0.611$_{\pm \text{0.000}}$ & & 0.331$_{\pm \text{0.110}}$ & 0.268$_{\pm \text{0.059}}$ \\
        GCNII & 0.465$_{\pm \text{0.002}}$ & 0.559$_{\pm \text{0.009}}$ & & 0.392$_{\pm \text{0.022}}$ & 0.473$_{\pm \text{0.061}}$ \\
        \midrule
        DAN & 0.338$_{\pm \text{0.005}}$ & 0.421$_{\pm \text{0.060}}$ & & 0.407$_{\pm \text{0.015}}$ & 0.422$_{\pm \text{0.015}}$ \\
        DANN & 0.368$_{\pm \text{0.021}}$ & 0.381$_{\pm \text{0.013}}$ & & 0.409$_{\pm \text{0.019}}$ & 0.419$_{\pm \text{0.022}}$ \\
        MDD & 0.349$_{\pm \text{0.029}}$ & 0.391$_{\pm \text{0.034}}$ & & 0.388$_{\pm \text{0.012}}$ & 0.421$_{\pm \text{0.015}}$ \\ \midrule
        AdaGCN & 0.451$_{\pm \text{0.013}}$ & 0.566$_{\pm \text{0.042}}$ & & 0.498$_{\pm \text{0.057}}$ & 0.516$_{\pm \text{0.025}}$  \\
        UDA-GCN &\bf 0.516$_{\pm \text{0.028}}$ & 0.600$_{\pm \text{0.014}}$ & & 0.471$_{\pm \text{0.010}}$ & 0.468$_{\pm \text{0.009}}$ \\
        EGI & 0.489$_{\pm \text{0.036}}$ & 0.404$_{\pm \text{0.006}}$ & & 0.494$_{\pm \text{0.030}}$ & 0.516$_{\pm \text{0.010}}$ \\ \midrule
        {\bf \model-N} & 0.475$_{\pm \text{0.011}}$ &\bf 0.635$_{\pm \text{0.009}}$ & &\bf 0.567$_{\pm \text{0.042}}$ &\bf 0.541$_{\pm \text{0.008}}$ \\
    \bottomrule
    \end{tabular}
    \caption{Cross-network node classification on the citation and social networks}
    \label{tab:citation_classification}
\end{table}

\begin{table}[!t]
    \centering
    \small
    \setlength\tabcolsep{3.7pt}
    \begin{tabular}{|l|ccccc|}
    \toprule
        \multirow{2}{*}{Methods} & \multicolumn{2}{c}{M $\rightarrow$ MU} & & \multicolumn{2}{c|}{MU $\rightarrow$ M} \\ \cline{2-3}\cline{5-6}
         & MAE & $R^2$ && MAE & $R^2$ \\\midrule
        GCN & 0.489$_{\pm \text{0.005}}$	&0.132$_{\pm \text{0.020}}$&&	0.679$_{\pm \text{0.030}}$&	0.295$_{\pm \text{0.051}}$ \\
        GCNII & 0.467$_{\pm \text{0.022}}$&	0.192$_{\pm \text{0.056}}$&&	0.687$_{\pm \text{0.042}}$ &	0.254$_{\pm \text{0.023}}$ \\
        DANN & 0.492$_{\pm \text{0.002}}$	&0.104$_{\pm \text{0.010}}$&&	0.721$_{\pm \text{0.030}}$&	0.192$_{\pm \text{0.069}}$\\
        RSD & 0.523$_{\pm \text{0.011}}$ &	0.019$_{\pm \text{0.061}}$ &&	0.729$_{\pm \text{0.009}}$&	0.170$_{\pm \text{0.047}}$ \\
        AdaGCN & 0.454$_{\pm \text{0.069}}$&	0.245$_{\pm \text{0.082}}$&&\bf	0.649$_{\pm \text{0.023}}$&	0.351$_{\pm \text{0.038}}$ \\
        UDA-GCN~ & 0.449$_{\pm \text{0.072}}$&	0.257$_{\pm \text{0.086}}$&&	0.684$_{\pm \text{0.012}}$&	0.271$_{\pm \text{0.003}}$ \\
        {\bf \model-N} &\bf 0.353$_{\pm \text{0.038}}$ &\bf	0.527$_{\pm \text{0.092}}$&&	0.652$_{\pm \text{0.007}}$&\bf	0.352$_{\pm \text{0.031}}$ \\
    \bottomrule
    \end{tabular}
    \caption{Plant phenotyping on the agriculture data set}
    \label{tab:agriculture}
\end{table}

\begin{table*}[!t]
\centering
\small
\setlength\tabcolsep{4pt}
\begin{tabular}{|l|ccccccccccc|}
\toprule
\multirow{2}{*}{Methods} & \multicolumn{3}{c}{CD $\rightarrow$ Music} &  & \multicolumn{3}{c}{Music $\rightarrow$ CD} &  & \multicolumn{3}{c|}{Book $\rightarrow$ Movie} \\ \cline{2-4}\cline{6-8}\cline{10-12}
                         & HR@10       & MRR@10      & NDCG@10      &  & HR@10       & MRR@10      & NDCG@10      &  & HR@10       & MRR@10      & NDCG@10      \\\midrule
BPRMF & 0.182$_{\pm \text{0.003}}$  & 0.061$_{\pm \text{0.003}}$  & 0.089$_{\pm \text{0.003}}$  & & 0.259$_{\pm \text{0.008}}$  & 0.097$_{\pm \text{0.006}}$  & 0.134$_{\pm \text{0.007}}$  & & 0.198$_{\pm \text{0.003}}$  & 0.070$_{\pm \text{0.002}}$  & 0.099$_{\pm \text{0.002}}$  \\
NeuMF & 0.286$_{\pm \text{0.012}}$ & 0.104$_{\pm \text{0.007}}$ & 0.145$_{\pm \text{0.008}}$ & & 0.328$_{\pm \text{0.014}}$ & 0.109$_{\pm \text{0.015}}$ & 0.160$_{\pm \text{0.013}}$ & & 0.294$_{\pm \text{0.020}}$ & 0.102$_{\pm \text{0.006}}$ & 0.146$_{\pm \text{0.009}}$ \\\midrule
CoNet & 0.405$_{\pm \text{0.018}}$ & 0.161$_{\pm \text{0.002}}$ & 0.214$_{\pm \text{0.006}}$ & & 0.333$_{\pm \text{0.010}}$ & 0.119$_{\pm \text{0.018}}$ & 0.168$_{\pm \text{0.015}}$ & & 0.319$_{\pm \text{0.023}}$ & 0.116$_{\pm \text{0.020}}$ & 0.162$_{\pm \text{0.020}}$ \\
CGN & 0.357$_{\pm \text{0.018}}$ & 0.120$_{\pm \text{0.015}}$ & 0.175$_{\pm \text{0.015}}$ & & 0.476$_{\pm \text{0.042}}$ & 0.192$_{\pm \text{0.020}}$ & 0.255$_{\pm \text{0.026}}$ & & 0.359$_{\pm \text{0.032}}$ & 0.136$_{\pm \text{0.018}}$ & 0.187$_{\pm \text{0.021}}$ \\ 
PPGN & 0.419$_{\pm \text{0.016}}$ &	0.179$_{\pm \text{0.008}}$ &	0.231$_{\pm \text{0.009}}$ & & 0.564$_{\pm \text{0.044}}$ & 0.278$_{\pm \text{0.032}}$ & 0.336$_{\pm \text{0.035}}$ & & 0.489$_{\pm \text{0.011}}$ & 0.239$_{\pm \text{0.007}}$ & 0.291$_{\pm \text{0.007}}$ \\
EGI & 0.446$_{\pm \text{0.011}}$ & 0.196$_{\pm \text{0.003}}$ & 0.250$_{\pm \text{0.006}}$ & & 0.599$_{\pm \text{0.007}}$ & 0.265$_{\pm \text{0.015}}$ & 0.338$_{\pm \text{0.009}}$ & & 0.461$_{\pm \text{0.021}}$ & 0.224$_{\pm \text{0.017}}$ & 0.274$_{\pm \text{0.014}}$ \\ \midrule
{\bf \model-R} &\bf 0.450$_{\pm \text{0.006}}$ &\bf 0.197$_{\pm \text{0.002}}$ &\bf 0.251$_{\pm \text{0.003}}$ & &\bf 0.600$_{\pm \text{0.011}}$ &\bf 0.313$_{\pm \text{0.007}}$ &\bf 0.373$_{\pm \text{0.008}}$ & &\bf 0.505$_{\pm \text{0.022}}$ &\bf 0.249$_{\pm \text{0.004}}$ &\bf 0.302$_{\pm \text{0.008}}$ \\
\bottomrule
\end{tabular}
\caption{Cross-domain recommendation on Amazon data set with overlapping users}
\label{tab:cdr}
\end{table*}

\begin{table*}[!t]
\centering
\small
\setlength\tabcolsep{3.5pt}
\begin{tabular}{|l|ccccccccccc|}
\toprule
\multirow{2}{*}{Methods} & \multicolumn{3}{c}{CD $\rightarrow$ Music} &  & \multicolumn{3}{c}{Music $\rightarrow$ CD} &  & \multicolumn{3}{c|}{Book $\rightarrow$ Movie} \\ \cline{2-4}\cline{6-8}\cline{10-12}
                         & HR@10       & MRR@10      & NDCG@10      &  & HR@10       & MRR@10      & NDCG@10      &  & HR@10       & MRR@10      & NDCG@10     \\\midrule
BPRMF & 0.167$_{\pm \text{0.006}}$ & 0.058$_{\pm \text{0.002}}$ & 0.083$_{\pm \text{0.003}}$ && 0.122$_{\pm \text{0.001}}$ & 0.039$_{\pm \text{0.001}}$ & 0.058$_{\pm \text{0.001}}$ && 0.165$_{\pm \text{0.004}}$ & 0.056$_{\pm \text{0.002}}$ & 0.081$_{\pm \text{0.002}}$  \\ 
NeuMF & 0.283$_{\pm \text{0.020}}$ & 0.101$_{\pm \text{0.012}}$ & 0.143$_{\pm \text{0.014}}$ && 0.133$_{\pm \text{0.006}}$ & 0.042$_{\pm \text{0.002}}$ & 0.063$_{\pm \text{0.003}}$ && 0.233$_{\pm \text{0.006}}$ & 0.083$_{\pm \text{0.002}}$ & 0.117$_{\pm \text{0.002}}$  \\ \midrule
NeuMF (S+T) & 0.312$_{\pm \text{0.012}}$ & 0.123$_{\pm \text{0.010}}$ & 0.165$_{\pm \text{0.009}}$ && 0.163$_{\pm \text{0.016}}$ & 0.056$_{\pm \text{0.009}}$ & 0.081$_{\pm \text{0.010}}$ && 0.271$_{\pm \text{0.015}}$ & 0.110$_{\pm \text{0.012}}$ & 0.146$_{\pm \text{0.013}}$ \\
EGI & 0.415$_{\pm \text{0.008}}$ & 0.169$_{\pm \text{0.028}}$ & 0.224$_{\pm \text{0.023}}$ &&\bf 0.231$_{\pm \text{0.035}}$ & 0.099$_{\pm \text{0.011}}$ & 0.128$_{\pm \text{0.015}}$ && 0.252$_{\pm \text{0.002}}$ & 0.145$_{\pm \text{0.004}}$ & 0.167$_{\pm \text{0.004}}$  \\ \midrule
{\bf \model-R} &\bf 0.423$_{\pm \text{0.001}}$ &\bf 0.185$_{\pm \text{0.001}}$ &\bf 0.238$_{\pm \text{0.000}}$ && 0.222$_{\pm \text{0.027}}$ &\bf 0.110$_{\pm \text{0.004}}$ &\bf 0.135$_{\pm \text{0.008}}$ &&\bf 0.324$_{\pm \text{0.059}}$ &\bf 0.187$_{\pm \text{0.022}}$ &\bf 0.216$_{\pm \text{0.029}}$ \\
\bottomrule
\end{tabular}
\caption{Cross-domain recommendation on Amazon data set with disjoint users}
\label{tab:cdr_disjoint}
\end{table*}

\subsection{Experimental Setup}
\subsubsection{Data Sets}
For cross-network node classification, we use the following data sets: Airport networks~\cite{ribeiro2017struc2vec} (Brazil, USA and Europe); Citation network~\cite{wu2020unsupervised,tang2008arnetminer} (ACMv9 (A) and DBLPv8 (D)); Social network~\cite{shen2020adversarial,li2015unsupervised} (Blog1 (B1) and Blog2 (B2)); and Agriculture data~\cite{wang2021unique} (Maize (M) and Maize\_UNL (MU)). 

For cross-domain recommendation, we evaluate the models on the Amazon data set~\cite{he2016ups}. We adopt two pairs of real-world cross-domain data sets from Amazon-5cores, including CD and Music, Book and Movie. Note that most existing cross-domain recommendation algorithms~\cite{hu2018conet,zhang2020learning} assume that source and target domains have the same group of users. To validate the effectiveness of our proposed approach, we consider two scenarios: (1) Overlapping users: following~\cite{hu2018conet}, source and target domains have the same group of users; (2) Disjoint users: the users of source and target domains are different.

\subsubsection{Baselines}
For cross-network node classification, we use the following baselines: (1) SourceOnly: GCN~\cite{kipf2017semi}, SGC~\cite{wu2019simplifying}, GCNII~\cite{chen2020simple}; (2) Feature-only adaptation: DAN~\cite{long2015learning}, DANN~\cite{ganin2016domain}, MDD~\cite{zhang2019bridging}; (3) Cross-network adaptation: AdaGCN~\cite{dai2022graph}, UDAGCN~\cite{wu2020unsupervised}, EGI~\cite{zhu2021transfer}.

For cross-domain recommendation, we use the following baselines: (1) Single-domain recommendation: BPR~\cite{rendle2009bpr}, NeuMF~\cite{he2017neural}; (2) Cross-domain recommendation: CoNet~\cite{hu2018conet}, PPGN~\cite{zhao2019cross}, CGN~\cite{zhang2020learning}, EGI~\cite{zhu2021transfer}.

\subsubsection{Model Configuration}
We adopt two hidden layers in the base GCN model when implementing the \model\footnote{\tt \url{https://github.com/jwu4sml/GRADE}} algorithms. We set $\lambda=0.02$ for cross-network node classification and $\lambda=0.1$ for the cross-domain recommendation.

\subsection{Cross-Network Node Classification}
Table~\ref{tab:airport_classification} and Table~\ref{tab:citation_classification} provide the cross-network node classification results of {\bf \model-N}. Here we report the classification accuracy, i.e., mean and standard deviation over 5 runs. We observe that (1) the cross-network node classification algorithms outperform the vanilla graph neural networks and the feature-only adaptation methods, and (2) in most cases, the proposed {\bf \model-N} algorithm improves the node classification performance (up to 13\%) over baselines.

In addition, we investigate the effectiveness of {\bf \model-N} on the regression task. In this case, we use mean square error (MSE) as the loss function of Eq. (\ref{eq:grade_nc}). Table~\ref{tab:agriculture} provides the results of {\bf \model-N} on the agriculture data set. Here we use two regression evaluation metrics: mean absolute error (MAE) and $R^2$. Since MDD~\cite{zhang2019bridging} focuses only on the classification setting, we use another state-of-the-art adaptation regression baseline RSD~\cite{chen2021representation} in our experiments. It is observed that our proposed {\bf \model-N} outperforms the state-of-the-art baselines for both MAE (lower is better) and $R^2$ (higher is better).

\subsection{Cross-Domain Recommendation}
\subsubsection{Results on overlapping users}
Table~\ref{tab:cdr} provides the cross-domain recommendation results on the Amazon data set. Here we use three recommendation metrics to evaluate our algorithms: hit ratio (HR@$k$), mean reciprocal rank (MRR@$k$), and normalized discounted cumulative gain (NDCG@$k$) where $k$ is 10. Following~\cite{hu2018conet,zhang2020learning}, we only consider the users shared by both source and target domains. We have the following observations. (1) Single-domain recommendation methods study the user preference in the target domain from limited observed user-item interactions. They have inferior performance due to network sparsity. (2) Cross-domain recommendation methods improve the model performance by leveraging the user preference information from the source domain. (3) The proposed {\bf \model-R} outperforms the state-of-the-art cross-domain recommendation baselines.

\subsubsection{Results on disjoint users}
Table~\ref{tab:cdr_disjoint} provides the results when the users of source and target domains are disjoint. Most existing cross-domain recommendation approaches~\cite{hu2018conet,zhang2020learning} cannot be applied to this scenario, because they require the shared users to explore common knowledge across domains. Thus, we only consider the single-domain recommendation baselines BPRMF~\cite{rendle2009bpr}, NeuMF~\cite{he2017neural}, and cross-domain recommendation baseline EGI~\cite{zhu2021transfer}. In addition, we also extend NeuMF to the cross-domain recommendation scenarios by incorporating the recommendation loss in the source domain (denoted as ``NeuMF (S+T)"). It is observed that the proposed {\bf \model-R} outperforms the baselines by a large margin (up to 18\% on HR@10) when the users are disjoint.

\begin{figure}[!t]
\centering
\subfigure[Base discrepancy]{\label{fig:a}\includegraphics[width=.23\textwidth]{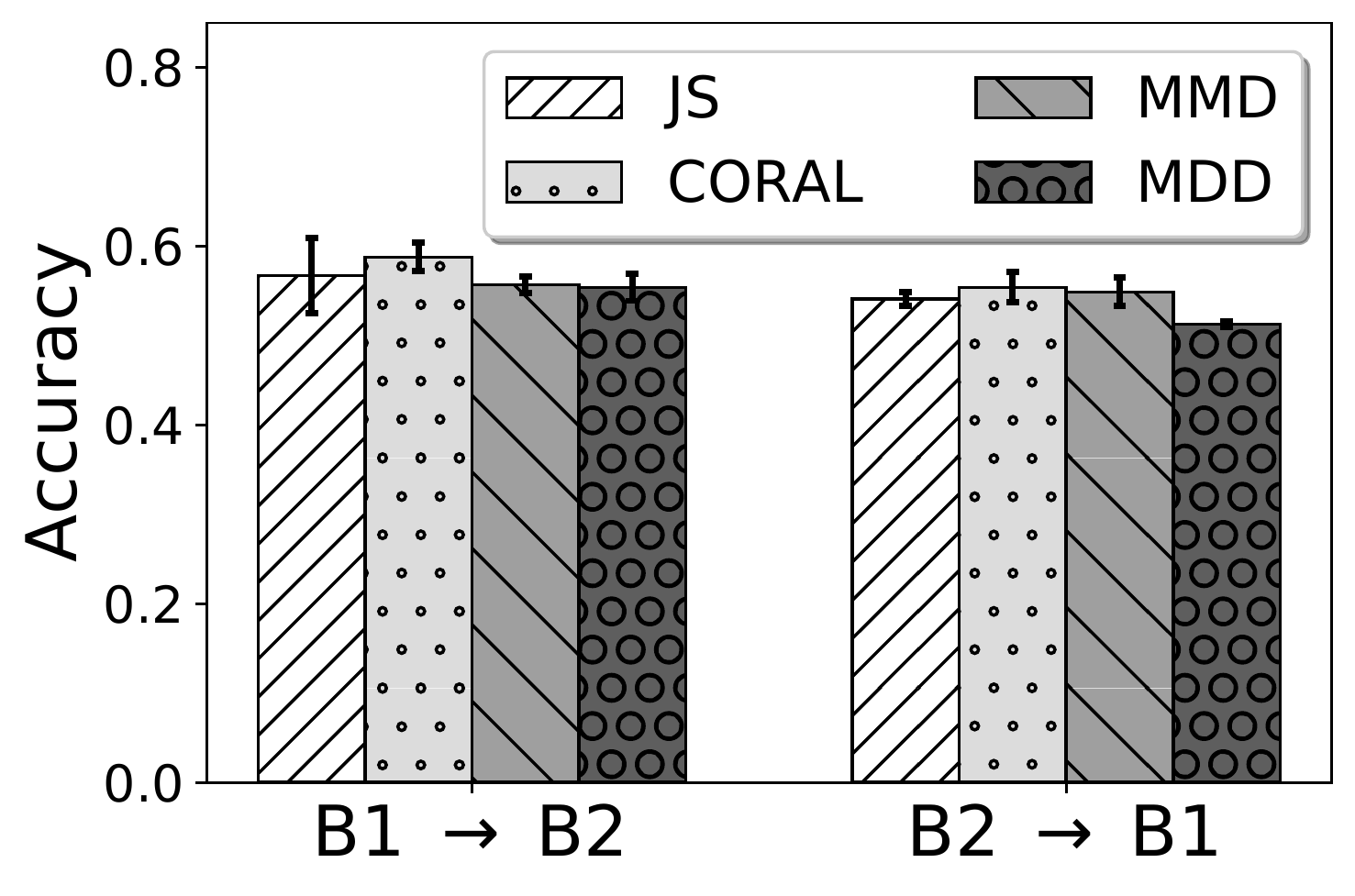}}
\subfigure[Base GNN]{\label{fig:b}\includegraphics[width=.23\textwidth]{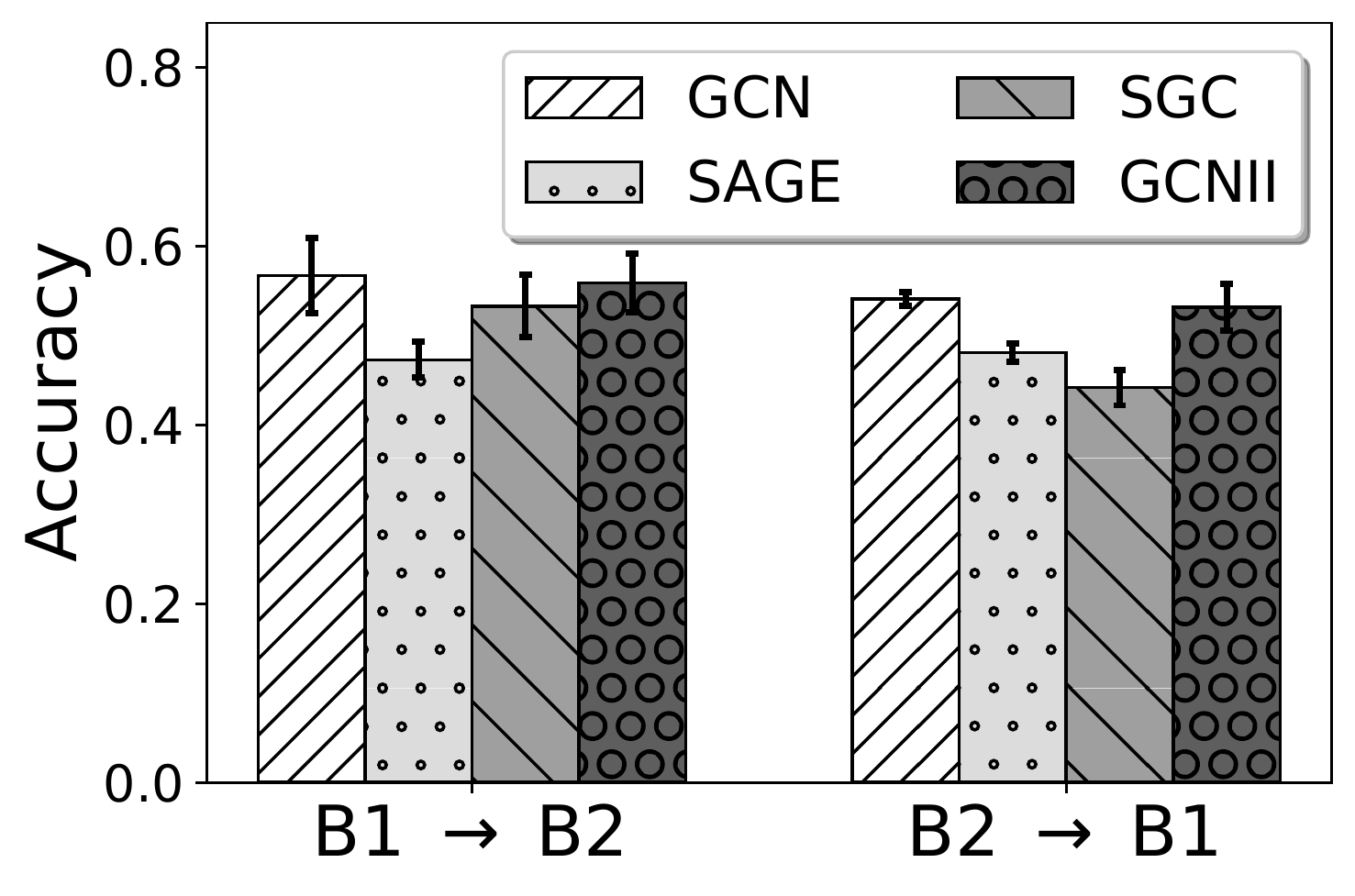}}
\caption{Performance of GRADE-N with different base discrepancies and base GNNs on social network}\label{fig:base_comparison}
\end{figure}

\subsection{Analysis}\label{sec:analysis}
\subsubsection{Flexibility}
Figure~\ref{fig:base_comparison} shows the performance of {\bf \model-N} with different base discrepancies and base graph neural network architectures on social networks. It shows that our \model\ framework is flexible to incorporate existing domain discrepancy measures (i.e., JS-divergence~\cite{ganin2016domain}, CORAL~\cite{sun2016deep}, MMD~\cite{gretton2012kernel} and MDD~\cite{zhang2019bridging}) and message-passing graph neural networks (i.e., GCN~\cite{kipf2017semi}, SAGE~\cite{hamilton2017inductive}, SGC~\cite{wu2019simplifying} and GCNII~\cite{chen2020simple}).

\subsubsection{Hyper-parameter Sensitivity}
\begin{figure}[!t]
\centering
\subfigure[Hyper-parameter sensitivity]{\label{fig:bsensitivity}\includegraphics[width=.23\textwidth]{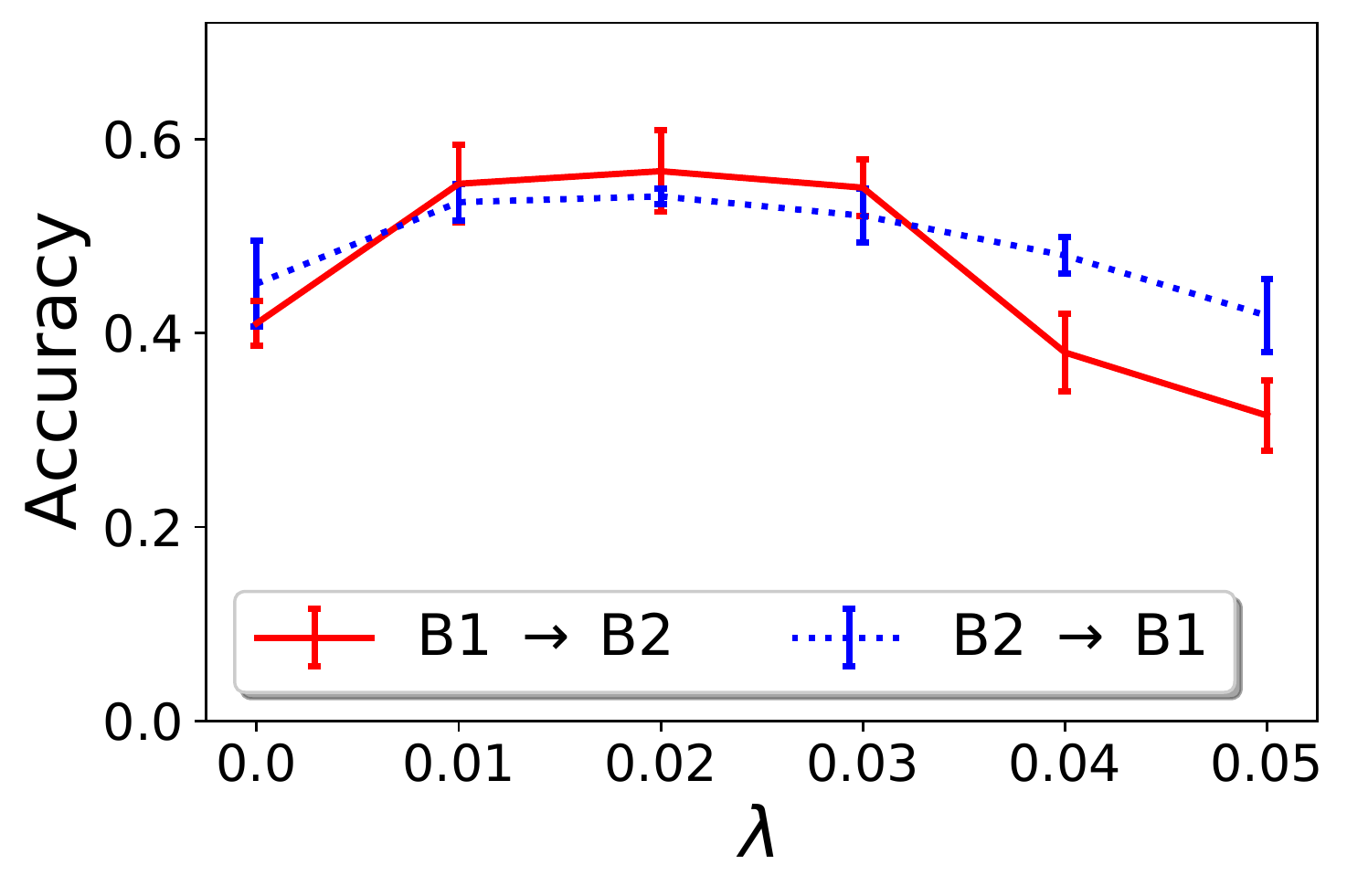}}
\subfigure[Efficiency analysis]{\label{fig:running_time}\includegraphics[width=.23\textwidth]{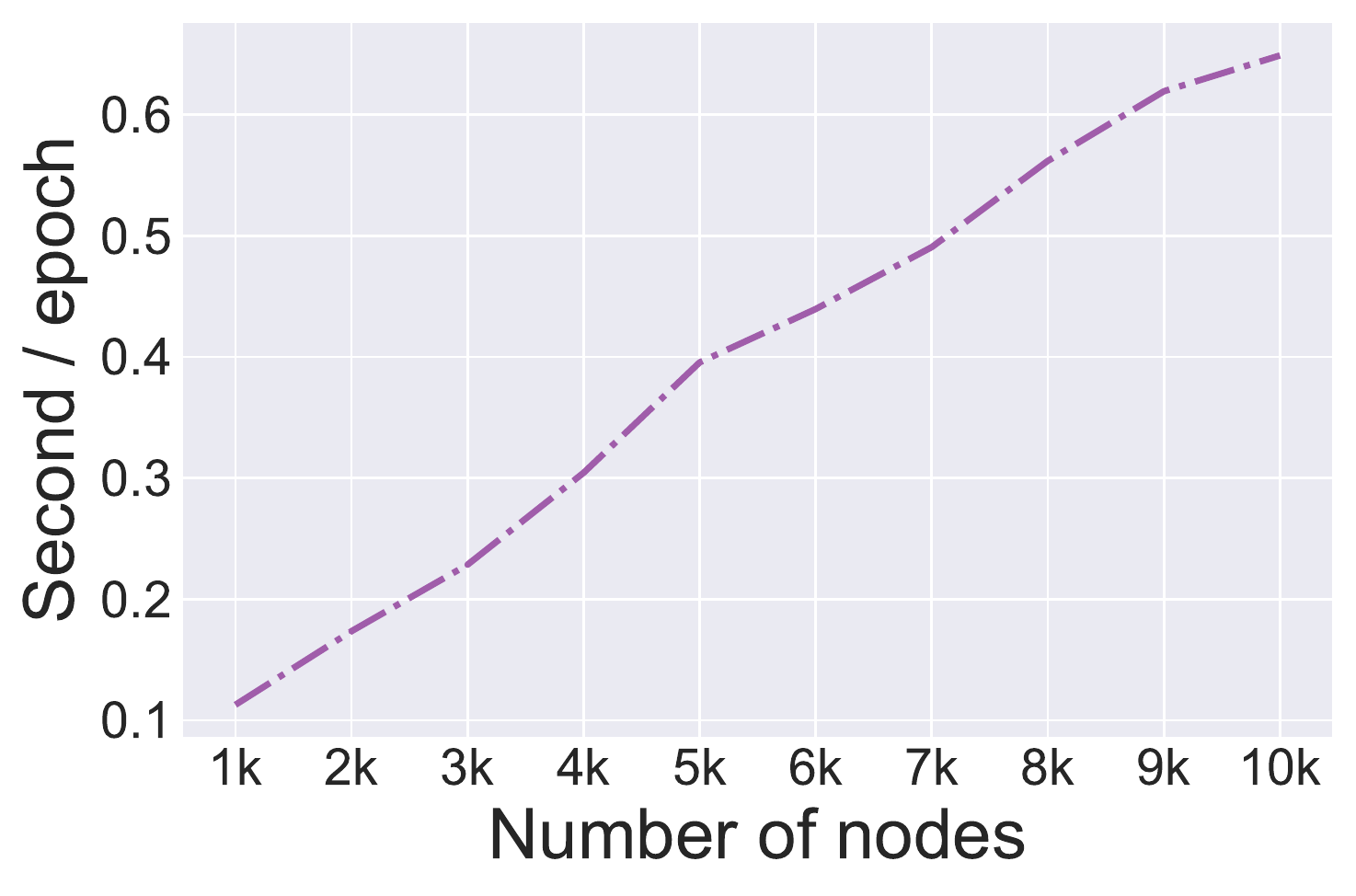}}
\caption{Model analysis of GRADE-N}\label{fig:analysis}
\end{figure}

We investigate the impact of $\lambda$ on {\bf \model-N}. Figure~\ref{fig:bsensitivity} shows the results on the social networks. It shows that the proposed {\bf \model-N} can achieve much better performance for $\lambda\in (0.01, 0.03)$. Thus, we use $\lambda = 0.02$ in the experiments.

\subsubsection{Computational Efficiency}
We investigate the computational efficiency of \model\ framework. Following~\cite{kipf2017semi}, we report the running time (measured in seconds wall-clock time) per epoch on the synthetic source and target graphs. Both graphs have $n$ (i.e., $n_s=n_t=n$) nodes and $2n$ edges.  As shown in Figure~\ref{fig:running_time}, we observe that the wall-clock time of {\bf \model-N} is linear with respect to the number of nodes within the source and target graphs.

\section{Conclusion}\label{sec:conclusion}
In this paper, we study the problem of cross-network transfer learning. We start by providing the theoretical analysis of cross-network transfer learning based on Graph Subtree Discrepancy. Then we propose a novel \model\ framework for cross-network transfer learning. Experiments demonstrate the effectiveness and efficiency of our \model\ framework.


\section*{Acknowledgements}
This work is supported by National Science Foundation under Award No. IIS-1947203, IIS-2117902, IIS-2137468, and Agriculture and Food Research Initiative (AFRI) grant no. 2020-67021-32799/project accession no.1024178 from the USDA National Institute of Food and Agriculture. The views and conclusions are those of the authors and should not be interpreted as representing the official policies of the funding agencies or the government.

\bibliography{aaai23}

\clearpage
\clearpage
\appendix
\onecolumn
\section{Appendix}

\subsection{More Discussion on Graph Subtree Discrepancy}\label{sec:properties}

\subsubsection{Properties of GSD}
we show that Graph Subtree Discrepancy (GSD) satisfies the following properties
\begin{lemma}\label{lem:GSD_metric}
If the ``relabeling'' function $f_m$ ($m=0,1,\cdots$) is injective, then for any graphs $G^s, G^t$, it holds
\begin{enumerate}[label=(\alph*)]
    \item $d_{GSD}\left(G^s, G^t\right)$ exists when $M$ goes to infinity.
    \item $d_b(G_{m-1}^s, G_{m-1}^t)\leq d_b(G_m^s, G_m^t)$ for any depth $m$.
    \item $d_{GSD}(G^s, G^t) = d_b(G^s, G^t)$, when $G^s, G^t$ are null graphs containing only isolated nodes.
    \item $d_{GSD}\left(G^s, G^t\right)$ is equivalent to the WL subtree kernel (see Definition~\ref{def:WL_kernel}), when $d_b(G_h^s, G_h^t) = \langle \phi(G_m^s), \phi(G_m^t) \rangle$ where $\phi(\cdot)$ counts the number of occurrences of subtree patterns.
\end{enumerate}
\end{lemma}

\begin{proof}
We first show the second {\bf property (b) of GSD}. The function $f_m(v)=f_m\left(f_{m-1}(v); \cup_{u\in N(v)} f_{m-1}(u) \right)$ learns the WL subtree of depth $m$ rooted at $v\in V$ by compressing the representation of $v$ and its neighbors from previous depth $m-1$. In this case, for any node $v^s \in V^s$ and $v^t \in V^t$, there are two scenarios for studying $f_m(v^s)$ and $f_m(v^t)$. (i) If $f_{m-1}(v^s) = f_{m-1}(v^t)$ at depth $m-1$, then $f_{m}(v^s) = f_{m}(v^t)$ only when $v^s$ and $v^t$ has the same node degree and their neighbors have the same representation; $f_{m}(v^s) \neq f_{m}(v^t)$, otherwise. (ii) If $f_{m-1}(v^s) \neq f_{m-1}(v^t)$ at depth $m-1$, then $f_{m}(v^s) \neq f_{m}(v^t)$. As a result, it can be seen that the distribution discrepancy of $G^s_m$ and $G^t_m$ would become larger than that of $G^s_{m-1}$ and $G^t_{m-1}$, as the WL subtrees become more diverse with the increase of depth $m$.
Therefore, we conclude that $d_b(G_m^s, G_m^t)$ is monotonically increasing with respect to the depth $m$, i.e., $d_b(G^s_{m-1}, G^t_{m-1})\leq d_b(G^s_m, G^t_m)$ for any integer $m$. 

For {\bf property (a) of GSD}, we need to show that $d_{GSD}\left(G^s, G^t\right)$ is bounded and monotonically increasing with respect to the depth $M$. 

For {\bf property (c) of GSD}, when one graph only contains the isolated nodes, it can be viewed as a set of independent nodes (variables). Thus, it would degenerate into a standard adaptation setting, where the distribution shift across domains can be measured by $d_b(G^s, G^t)$ on IID nodes. In other words, when there is no edge in the graph, each subtree at any depth $m$ rooted at $v$ is the node $v$ itself. Therefore, it holds that $d_b(G_0^s, G_0^t) =\cdots = d_b(G_m^s, G_m^t)$ for any $m$. Then we have $d_{GSD}(G^s, G^t) = d_b(G^s_0, G^t_0) = d_b(G^s, G^t)$.

For {\bf property (d) of GSD}, it can be derived using the definition of WL subtree kernel.
\end{proof}

\subsubsection{Benefits of Graph Subtree Discrepancy}
The benefits of Graph Subtree Discrepancy could be summarized as follows. (1) {\em Interpretability:} It measures the difference of subtree distributions in the source and target graphs. When the subtree representation is countable, it would be equivalent to the standard Weisfeiler-Leman graph subtree kernel by counting the number of occurrences of subtree patterns. (2) {\em Computational Efficiency:} When using GCN~\cite{kipf2017semi} as the base model for subtree representation learning and multi-layer perceptrons as the hypothesis space $\mathcal{H}$ for learning the subtree representation, the time complexity of GSD is $\mathcal{O}(M(n_s + n_t)d^2 + M(m_s + m_t)d)$ where $n_s, n_t$ denote the number of nodes within the source and target graphs, $m_s, m_t$ denote the number of edges within the source and target graphs, $d$ is the dimensionality of hidden layers, and $M$ is the number of neural layers.
(3) {\em Flexibility:} It enables the estimate of distribution shift across graphs using existing expressive graph neural networks~\cite{kipf2017semi,hamilton2017inductive,velivckovic2018graph} and standard domain discrepancy measures~\cite{gretton2012kernel,ganin2016domain,zhang2019bridging} (see Subsection~\ref{sec:analysis} for empirical analysis).

\subsubsection{Improved Graph Subtree Discrepancies}\label{sec:improved_disc}
We provide two improved variants of Graph Subtree Discrepancy by incorporating specific structure and label information.
\begin{itemize}
    \item {\tt Structure-aware GSD:} As one of the most important properties in a graph, node degree encodes the intrinsic graph structure~\cite{demonet_kdd19,tang2020investigating,geerts2021let} despite its simplicity. Following the Weisfeiler-Lehman graph isomorphism test~\cite{weisfeiler1968reduction,shervashidze2011weisfeiler}, two graphs can be easily distinguished from the node degree distribution. Therefore, we can design a degree-specific GSD by taking the node degree into consideration.
    \begin{equation}
        d_{GSD}^{deg}\left(G^s, G^t\right) = \lim_{M\to \infty} \frac{1}{M+1} \sum_{m=0}^M d_b\left(G_m^s\circ D^s, G_m^t\circ D^t\right)
    \end{equation}
    where $\circ$ denotes the vector concatenation of subtree representation around $v\in V$ and one-hot degree representation of $v$. This improves the quality of GSD, when the distribution shift of graphs is largely induced by the node attributes. In this case, source and target nodes are more likely to have the identical class label, when they have the same node degree. Thus, the degree-specific distribution discrepancy measure can better characterize the distribution shift across graphs.

    \item {\tt Label-informed GSD:} In the context of cross-network node classification, the class labels of nodes can help refine the definition of distribution shift across graphs. As demonstrated in previous works~\cite{zhao2019learning}, minimizing the feature-based marginal distribution discrepancy cannot guarantee the success of the knowledge transfer across domains. Therefore, inspired by~\cite{wu2020continuous}, we proposed a label-informed GSD by considering the distribution shifts across graphs induced by both input features and output labels.
    \begin{equation}\label{eq:labeled_GSD}
        d_{GSD}^{cls}\left(G^s, G^t\right) = \lim_{M\to \infty} \frac{1}{M+1} \sum_{m=0}^M d_b\left(G_m^s\circ Y^s, G_m^t\circ Y^t\right)
    \end{equation}
    Here, $\circ$ denotes the vector concatenation of subtree representation around $v\in V$ and the one-hot class label representation of $v$. In practice, when the class labels of target nodes are unknown, we can use their pseudo-labels to estimate this discrepancy.
\end{itemize}
Similarly, we can derive the generalization error bounds based on the improved GSD, because it holds that $d_{GSD}\left(G^s, G^t\right) \leq d_{GSD}^{deg}\left(G^s, G^t\right)$ and $d_{GSD}\left(G^s, G^t\right) \leq d_{GSD}^{cls}\left(G^s, G^t\right)$.

\begin{figure}[h]
\centering
\subfigure[Label-informed discrepancy]{\label{fig:aadvanced_comparison}\includegraphics[width=.4\textwidth]{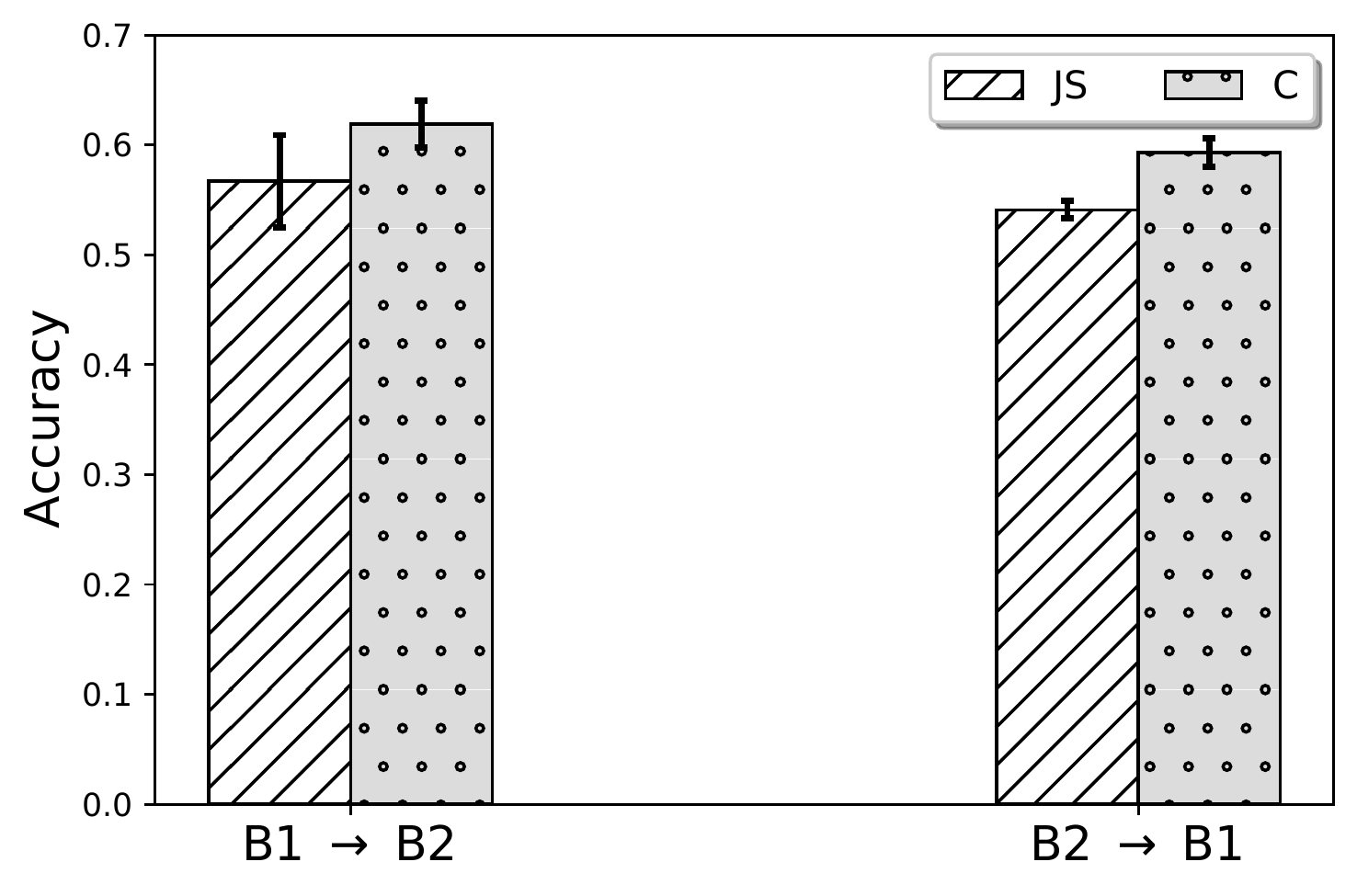}}
\subfigure[Degree-specific discrepancy]{\label{fig:badvanced_comparison}\includegraphics[width=.4\textwidth]{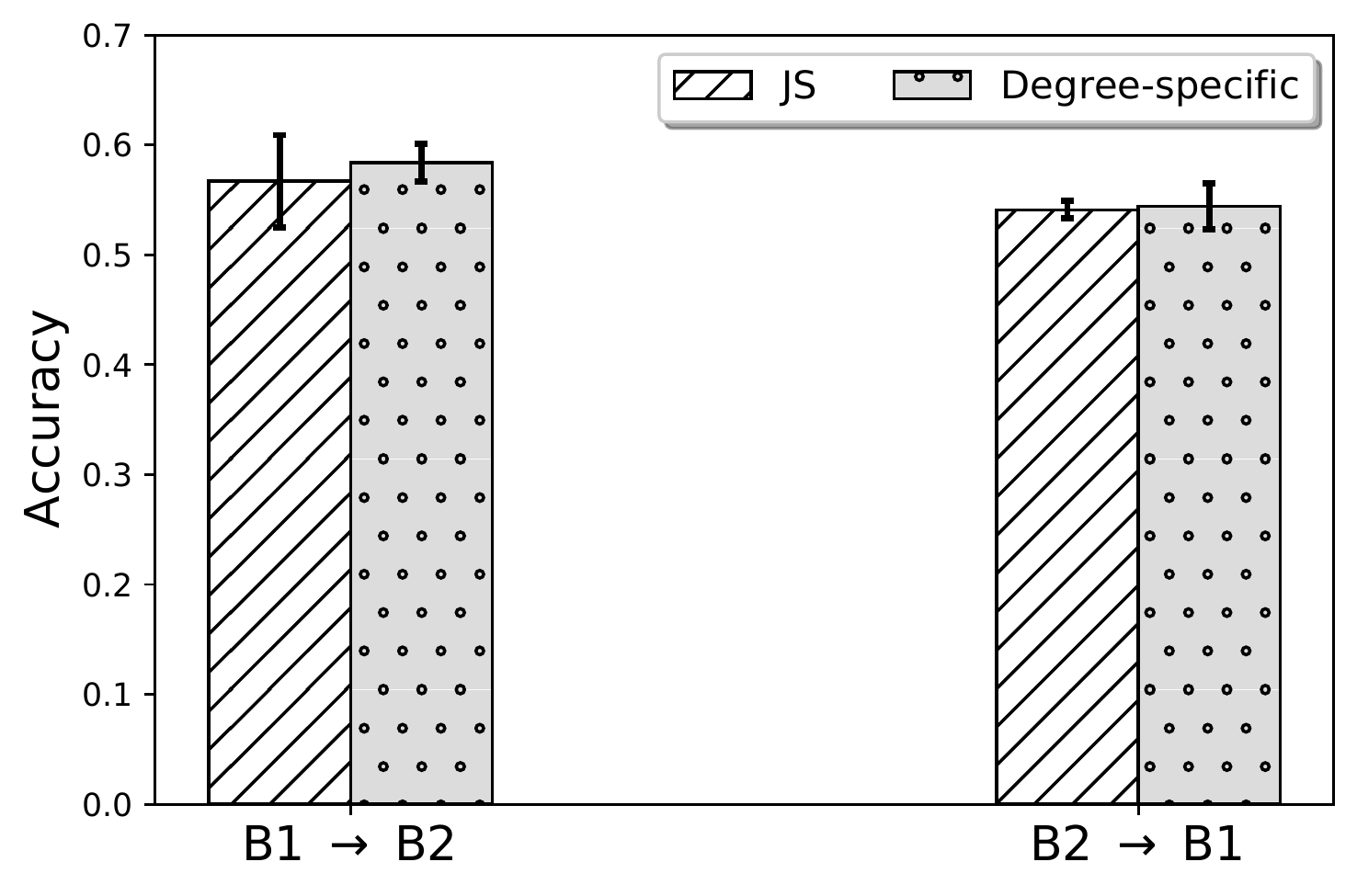}}
\vspace{-3mm}
\caption{Performance of {\bf \model-N} with improved discrepancies on social network}\label{fig:advanced_comparison}
\vspace{-2mm}
\end{figure}

Figure~\ref{fig:advanced_comparison} provides the results of {\bf \model-N} using the improved discrepancy above. Since the target domain is unlabeled, we use the pseudo-labels of target nodes to estimate the label-informed GSD of Eq. (\ref{eq:labeled_GSD}). It is observed that both label and degree information can help improve the quality of GSD, thus leading to better model performance.


\subsection{Illustration of \model\ Framework}
\begin{figure}[h]
    \centering
    \includegraphics[width=.7\textwidth]{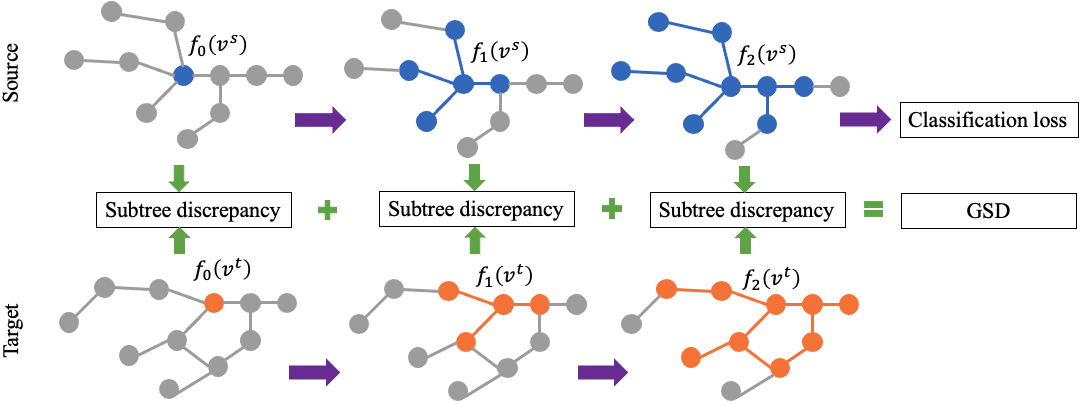}
    \caption{Illustration of \model\ on cross-network node classification}
    \label{fig:illustration_of_framework}
\end{figure}

Figure~\ref{fig:illustration_of_framework} shows the proposed \model\ framework on cross-network node classification, where Graph Subtree Discrepancy (GSD) is defined over the subtrees involving both the structure and node attributes in the center node’s local neighborhood.

\subsection{Proof of Theorem~\ref{thm:TL_node_classification}}
Theorem~\ref{thm:TL_node_classification} states that
assume there are a source graph $G^s$ and a target graph $G^t$ and the base domain discrepancy $d_b(\cdot,\cdot)$ of GSD is instantiated by the discrepancy distance (see Eq. (\ref{eq:discrepancy_distance})). Given a message-passing GNN with the feature extractor $f$ and the hypothesis $h\in \mathcal{H}$, the node classification error in the target graph can be bounded as follows.
\begin{equation*}
    \epsilon_t(h\circ f) \leq \epsilon_s(h\circ f) + d_{GSD}\left(G^s, G^t\right) + \lambda^* + R^*
\end{equation*}
where $\lambda^* = \mathbb{E}_{v\in V^t}[\mathcal{L}(h^s_*(f(v)), h^t_*(f(v)))]$ measures the prediction difference of optimal source and target hypotheses on the target nodes, and $R^*=\mathbb{E}_{v\in V^s}[\mathcal{L}(y, h^s_*(f(v)))] + \mathbb{E}_{v\in V^t}[\mathcal{L}(h^t_*(f(v)), y)]$ is the Bayes error on the source and target graphs. $y$ is the class label of $v$. In this case, $h^s_* \in \arg\min_{h\in \mathcal{H}}$ $\mathbb{E}_{v\in V^s}[\mathcal{L}(h(f(v)), y)]$ and $h^t_* \in \arg\min_{h\in \mathcal{H}} \mathbb{E}_{v\in V^t}[\mathcal{L}(h(f(v)), y)]$ are optimal source and target hypotheses, respectively.

\begin{proof}
Using Lemma~\ref{lem:GSD_metric}, it holds that for any $M>0$
\begin{align*}
    &\quad \frac{1}{M+1} \sum_{m=0}^M d_b\left(G_m^s, G_m^t\right) - \frac{M-L+1}{M+1} d_b\left(G_L^s, G_L^t\right) \\
    &= \frac{1}{M+1} \sum_{m=0}^{L-1} d_b\left(G_m^s, G_m^t\right) + \frac{1}{M+1} \sum_{m=L}^M \left( d_b\left(G_m^s, G_m^t\right) - d_b\left(G_L^s, G_L^t\right) \right) \geq 0
\end{align*}
Then it holds that $d_{GSD}\left(G^s, G^t\right) = \lim_{M\to \infty} \frac{1}{M+1} \sum_{m=0}^M d_b\left(G_m^s, G_m^t\right) \geq \lim_{M\to \infty} \frac{M-L+1}{M+1} d_b\left(G_L^s, G_L^t\right) = d_b\left(G_L^s, G_L^t\right)$. That means that the following holds.
\begin{align*}
    d_{GSD}\left(G^s, G^t\right) &= \lim_{M\to \infty} \frac{1}{M+1} \sum_{m=0}^M \sup_{h, h'\in {\mathcal{H}}} \Big| \mathbb{E}_{v\in V^s}\left[ \mathcal{L}\left(h\left(f_m(v)\right), h'\left(f_m(v)\right)\right) \right] - \mathbb{E}_{v\in V^t}\left[ \mathcal{L}\left(h\left(f_m(v)\right), h'\left(f_m(v)\right)\right) \right] \Big| \\
    &\geq \sup_{h, h'\in \Tilde{\mathcal{H}}} \big| \mathbb{E}_{v\in V^s}\left[ \mathcal{L}\left(h\left(f_L(v)\right), h'\left(f_L(v)\right)\right) \right] - \mathbb{E}_{v\in V^t}\left[ \mathcal{L}\left(h\left(f_L(v)\right), h'\left(f_L(v)\right)\right) \right] \big| \\
    &\geq \mathbb{E}_{v\in V^t}[\mathcal{L}(h(f_L(v)), h^s_*(f_L(v)))] - \mathbb{E}_{v\in V^s}[\mathcal{L}(h(f_L(v)), h^s_*(f_L(v)))] \\
    &= \mathbb{E}_{v\in V^t}[\mathcal{L}(h(f(v)), h^s_*(f(v)))] - \mathbb{E}_{v\in V^s}[\mathcal{L}(h(f(v)), h^s_*(f(v)))]
\end{align*}

If the classification error is defined using the output of the last graph convolutional layer, i.e., $\epsilon_t(h\circ f) = \mathbb{E}_{v\in G}[L(h(f_L(v)), y)]$, we have the following results.
\begin{align*}
    & \epsilon_t(h\circ f) = \mathbb{E}_{v\in V^t}[\mathcal{L}(h(f(v)), y)] \\
    &\leq \mathbb{E}_{v\in V^t}[\mathcal{L}(h(f(v)), h^t_*(f(v)))] + \mathbb{E}_{v\in V^t}[\mathcal{L}(h^t_*(f(v)), y)] \\
    &\leq \mathbb{E}_{v\in V^t}[\mathcal{L}(h(f(v)), h^s_*(f(v)))] + \mathbb{E}_{v\in V^t}[\mathcal{L}(h^s_*(f(v)), h^t_*(f(v)))] + \mathbb{E}_{v\in V^t}[\mathcal{L}(h^t_*(f(v)), y)] \\
    &\leq \mathbb{E}_{v\in V^s}[\mathcal{L}(h(f(v)), h^s_*(f(v)))] + d_{GSD}\left(G^s, G^t\right) + \mathbb{E}_{v\in V^t}[\mathcal{L}(h^s_*(f(v)), h^t_*(f(v)))] + \mathbb{E}_{v\in V^t}[\mathcal{L}(h^t_*(f(v)), y)] \\
    &\leq \epsilon_s(h\circ f) + \mathbb{E}_{v\in V^s}[\mathcal{L}(y, h^s_*(f(v)))] + d_{GSD}\left(G^s, G^t\right) + \mathbb{E}_{v\in V^t}[\mathcal{L}(h^s_*(f(v)), h^t_*(f(v)))] + \mathbb{E}_{v\in V^t}[\mathcal{L}(h^t_*(f(v)), y)]
\end{align*}
which completes the proof.
\end{proof}

\subsection{Corollary~\ref{cor:other_classification_NN}}\label{sec:other_classification_NN}
\begin{corollary}\label{cor:other_classification_NN}
With assumptions in Theorem~\ref{thm:TL_node_classification}, and let $\mathcal{L}(y,\Tilde{y}) = |y-\Tilde{y}|$ and the hypothesis class $\mathcal{H}$ is given by the multi-layer perceptrons, if the classification error is defined using the jumping knowledge, i.e., $\epsilon_t(h\circ f) = \mathbb{E}_{v\in G}[\mathcal{L}(h(f(v)), y)]$ where $f(v) = [f_0(v), \cdots, f_L(v)]$, the node classification error in the target graph can also be bounded as follows.
\begin{equation*}
    \epsilon_t(h\circ f) \leq \epsilon_s(h\circ f) + 2d_{GSD}\left(G^s, G^t\right) + \lambda^* + R^*
\end{equation*}
where $\lambda^* = \mathbb{E}_{v\in V^t}[\mathcal{L}(h^s_*(f(v)), h^t_*(f(v)))]$, and $R^*=\mathbb{E}_{v\in V^s}[\mathcal{L}(y, $ $h^s_*(f(v)))] + \mathbb{E}_{v\in V^t}[\mathcal{L}(h^t_*(f(v)), y)]$.
\end{corollary}

\begin{proof}
Using Lemma~\ref{lem:GSD_metric}, it holds that for any $M>0$
\begin{align*}
    &\quad \frac{1}{M+1} \sum_{m=0}^M d_b\left(G_m^s, G_m^t\right) - \sum_{l=0}^L \frac{M-l+1}{(L+1)(M+1)} d_b\left(G_l^s, G_l^t\right) \\
    &= \frac{1}{L+1} \sum_{l=0}^L \left( \frac{1}{M+1} \sum_{m=0}^M d_b\left(G_m^s, G_m^t\right) - \frac{M-l+1}{M+1} d_b\left(G_l^s, G_l^t\right) \right) \geq 0
\end{align*}
Then it holds $d_{GSD}\left(G^s, G^t\right) = \lim_{M\to \infty} \frac{1}{M+1} \sum_{m=0}^M d_b\left(G_m^s, G_m^t\right) \geq \lim_{M\to \infty} \sum_{l=0}^L \frac{M-l+1}{(L+1)(M+1)} d_b\left(G_l^s, G_l^t\right) = \frac{1}{L+1} \sum_{l=0}^L d_b\left(G_l^s, G_l^t\right)$.
If the classification error is defined using the jumping knowledge, i.e., $\epsilon_t(h\circ f) = \mathbb{E}_{v\in V}[\mathcal{L}(h(f(v)), y)]$ where $f(v) = [f_0(v), \cdots, f_M(v)]$, the hypothesis class $\mathcal{H}$ used in the classification and GSD would be different. So we denote the hypothesis class of GSD as $\Tilde{\mathcal{H}}$ in this case. If the hypothesis class $\mathcal{H}$ is instantiated by the multi-layer perceptrons (MLPs), then there exists  $h_0, h_1, \cdots, h_L\in \Tilde{\mathcal{H}}$ ($\Tilde{\mathcal{H}}$ can also be multi-layer perceptrons) such that for any hypothesis $h\in \mathcal{H}$, it could be represented as $h(f(v)) = h([f_0(v), \cdots, f_L(v)]) = \frac{1}{L+1} \sum_{l=0}^L h_l(f_l(v))$. That can be derived by decomposing the first-layer parameters of $h$ as follows.
\begin{align*}
    W_1 [f_0(v), \cdots, f_M(v)] + b_1 = (W_{10}f_0(v) + b_{10}) + \cdots + (W_{1M}f_M(v) + b_{1M})
\end{align*}
where the weight $W_1$ can be divided into $M$ parts according to its columns, and $b_1 = b_{10} + \cdots + b_{1M}$, and the constant scaling factor $\frac{1}{L+1}$ can be added, as $\frac{1}{L+1} h_l(\cdot) \in \Tilde{\mathcal{H}}$. Therefore, we have
\begin{align*}
    &\quad 2d_{GSD}\left(G^s, G^t\right) \\
    &= \lim_{M\to \infty} \frac{2}{M+1} \sum_{m=0}^M \sup_{\Tilde{h}, \Tilde{h'}\in \Tilde{\mathcal{H}}} \Big| \mathbb{E}_{v\in V^s}\left[ \mathcal{L}\left(\Tilde{h}\left(f_m(v)\right), \Tilde{h'}\left(f_m(v)\right)\right) \right]  - \mathbb{E}_{v\in V^t}\left[ \mathcal{L}\left(\Tilde{h}\left(f_m(v)\right), \Tilde{h'}\left(f_m(v)\right)\right) \right] \Big| \\
    &\geq \frac{2}{L+1} \sum_{l=0}^L \sup_{\Tilde{h}, \Tilde{h'}\in \Tilde{\mathcal{H}}} \Big| \mathbb{E}_{v\in V^s}\left[ \mathcal{L}\left(\Tilde{h}\left(f_l(v)\right), \Tilde{h'}\left(f_l(v)\right)\right) \right] - \mathbb{E}_{v\in V^t}\left[ \mathcal{L}\left(\Tilde{h}\left(f_l(v)\right), \Tilde{h'}\left(f_l(v)\right)\right) \right] \Big| \\
    &= \frac{2}{L+1} \sum_{l=0}^L \sup_{\Tilde{h}, \Tilde{h'}\in \Tilde{\mathcal{H}}} \left| \int_{V} \left( p^s(v) - p^t(v) \right) \mathcal{L}\left(\Tilde{h}\left(f_l(v)\right), \Tilde{h'}\left(f_l(v)\right)\right) dv \right| \\
    &\geq \frac{1}{L+1} \sum_{l=0}^L \sup_{\Tilde{h}, \Tilde{h'}\in \Tilde{\mathcal{H}}} \int_{V} \left| p^s(v) - p^t(v) \right| \left|\Tilde{h}(f_l(v)) - \Tilde{h'}(f_l(v)) \right| dv \\
    &\geq \frac{1}{L+1} \sum_{l=0}^L \int_{V} \left|p^s(v) - p^t(v)\right| \left|h_l(f_l(v)) - h^s_{*l}(f_l(v)) \right| dv \\
    &\geq \int_{V} \left|p^s(v) - p^t(v)\right| \left|h(f(v)) - h^s_*(f(v)) \right| dv \\
    &\geq \mathbb{E}_{v\in V^t}[\mathcal{L}(h(f(v)), h^s_*(f(v)))] - \mathbb{E}_{v\in V^s}[\mathcal{L}(h(f(v)), h^s_*(f(v)))]
\end{align*}
which completes the proof.
\end{proof}

\subsection{Proof of Theorem~\ref{thm:TL_link_prediction}}
Theorem~\ref{thm:TL_link_prediction} states that
with assumptions in Theorem~\ref{thm:TL_node_classification}, and let $\mathcal{L}(y,\Tilde{y}) = |y-\Tilde{y}|$ and the hypothesis class $\mathcal{H}$ is given by the multi-layer perceptrons, if the loss of the link prediction is defined as $\epsilon^{link}(h\circ f) = \mathbb{E}_{u,v\in V\times V}[\mathcal{L}(h([f_L(u) || f_L(v)]), y)]$, then the link prediction error in the target graph can be bounded as follows.
\begin{align*}
    \epsilon_t^{link}(h) \leq \epsilon_t^{link}(h) + d_{GSD}\left(G_s, G_t\right) + \lambda^*_{link} + R^*_{link}
\end{align*}
where $\lambda^*_{link} = \mathbb{E}_{(u,v)\in V^t\times V^t}[\mathcal{L}(h^s_*([f(u) || f(v)]), h^t_*([f(u) || f(v)]))]$ measures the difference of optimal source and target hypotheses on the target graph, and $R^*_{link}=\mathbb{E}_{(u,v)\in V^s\times V^s}[\mathcal{L}(y, h^s_*([f(u) || f(v)]))] + \mathbb{E}_{(u,v)\in V^t\times V^t}[\mathcal{L}(h^t_*([f(u) || f(v)]), y)]$ is the Bayes error. In this case, $h^s_* \in \arg\min_{h\in \mathcal{H}}$ $\mathbb{E}_{(u,v)\in V^s\times V^s}[\mathcal{L}(h([f(u) || f(v)]), y)]$, and $h^t_* \in \arg\min_{h\in \mathcal{H}} \mathbb{E}_{(u,v)\in V^t\times V^t}[\mathcal{L}(h([f(u) || f(v)]), y)]$ are optimal source and target hypothesises, respectively.

\begin{proof}
It can be shown using a similar method in Corollary~\ref{cor:other_classification_NN}. Here the output feature function $f(\cdot)$ would stack the representations from a pair of nodes, i.e., $f(u, v) = [f_L(u) || f_L(v)]$.
\end{proof}

\subsection{Data Description}\label{sec:data_description}
For cross-network node classification, we use the following benchmark data sets:
\begin{itemize}
    \item Airport network~\cite{ribeiro2017struc2vec}: It contains three airport networks from Brazil, USA and Europe. Each node corresponds to an airport and the edge indicates the existence of commercial flights between two airports. The class labels of nodes are assigned based on the level of activity measured by flights or people that passed the airports. Following~\cite{zhu2021transfer}, we use node degree one-hot encoding as the node feature.
    \item Citation network~\cite{wu2020unsupervised}: It has two networks ACMv9 (A) and DBLPv8 (D) from ArnetMine~\cite{tang2008arnetminer}. Each node is a paper and each edge indicates the citation between two papers. Each paper is associated with a 7537-dimensional feature vector extracted from paper content. Its class label indicates the research topics.
    \item Social network~\cite{shen2020adversarial}: Blog1 (B1) and Blog2 (B2) are two disjoint social networks extracted from BlogCatalog~\cite{li2015unsupervised}. Each node is a blogger and each edge indicates the friendship between two bloggers. The blogger is associated with an 8189-dimensional feature vector extracted from the blogger’s self-description, and its class label indicates the joining group.
    \item Agriculture data~\cite{wang2021unique}: Plant Phenotyping predicts diverse traits (e.g., Nitrogen) of plants related to the plants’ growth using leaf hyperspectral reflectance. Here we use the agriculture data from two domains: Maize (M) and Maize\_UNL (MU) (i.e., maize data are measured from different locations). In our case, the task is to predict the Nitrogen content of maize using the leaf hyperspectral reflectance. Specifically, each example can be represented as a 1901-dimensional feature vector, which characterizes the spectral wavelengths 500-2400 nm. Then we can adopt $k$-NN to build the graph for each domain. We use $k=5$ in our experiments.
\end{itemize}
For cross-domain recommendation, we evaluate the models on the Amazon data set~\cite{he2016ups}. We adopt two pairs of real-world cross-domain data sets from Amazon-5cores, including CD (i.e., CDs and Vinyl) and Music (i.e., Digital Music), Book (i.e., Books) and Movie (i.e., Movies and TV). Note that most of the existing cross-domain recommendation algorithms~\cite{hu2018conet,zhang2020learning} assume that source and target domains have the same group of users. To validate the effectiveness of our proposed approach, we would like to consider the following two scenarios: (1) Overlapping users: following~\cite{hu2018conet}, source and target domains have the same group of users; (2) Disjoint users: the users of source and target domains are not overlapping.

Table~\ref{tab:data_statistics} summarizes all the data sets used in the experiments. All the experiments are performed on a Windows machine with four 3.80GHz Intel Cores, 64GB RAM, and one NVIDIA Quadro RTX 5000 GPU.

\begin{table}[!t]
    \centering
    \begin{tabular}{|l|l|rrr|}
    \toprule
        \multicolumn{2}{|c|}{Data} & \#nodes & \#edges & \#classes \\ \midrule
        \multirow{3}{*}{Airport} & USA & 1,190 & 13,599 & 4 \\
        & Brazil & 131 & 1,038 & 4 \\
        & Europe & 399 & 5,995 & 4 \\ \midrule
        \multirow{2}{*}{Citation} & ACMv9 & 7,410 & 11,135 & 6 \\
        & DBLPv8 & 5,578 & 7,341 & 6 \\ \midrule
        \multirow{2}{*}{Social} & Blog1 & 2,300 & 33,471 & 6 \\
        & Blog2 & 2,896 & 53,836 & 6 \\ \midrule
        \multirow{2}{*}{Agriculture} & Maize & 349 & 1,745 & - \\
        & Maize\_UNL & 1,210 & 6,050 & - \\ \midrule \midrule
        \multicolumn{2}{|c|}{Data} & \#users & \#items & \#ratings \\ \midrule
        \multirow{2}{*}{CD vs. Music (overlapping)} & CD & 5,000 & 55,312 & 353,942 \\
        & Music & 5,000 & 90,248 & 155,192 \\ \midrule
        \multirow{2}{*}{Book vs. Movie (overlapping)} & Book & 5,000 & 90,248 & 155,192 \\
        & Movie & 5,000 & 28,796 & 100,302 \\ \midrule
        \multirow{2}{*}{CD vs. Music (disjoint)} & CD & 5,000 & 27,838 & 51,190 \\
        & Music & 5,000 & 3,568 & 58,408 \\ \midrule
        \multirow{2}{*}{Book vs. Movie (disjoint)} & Book & 5,000 & 51,968 & 72,804  \\
        & Movie & 5,000 & 21,070 & 52,376 \\
    \bottomrule
    \end{tabular}
    \caption{Data statistics}
    \label{tab:data_statistics}
\end{table}

\end{document}